\documentclass[a4paper]{article}

\usepackage{float}
\usepackage[
  a4paper,top=3cm,bottom=2cm,left=3cm,right=3cm,marginparwidth=1.75cm]{geometry}
\usepackage{amsmath}
\usepackage{amsthm,bm,mathtools,etoolbox,amsfonts,bbm}
\usepackage{hyperref}
\hypersetup{colorlinks,linkcolor={blue},citecolor={blue},urlcolor={blue}}    
\usepackage[nameinlink,capitalise]{cleveref}
\usepackage[utf8]{inputenc} 
\usepackage[T1]{fontenc}    
\usepackage{url} 
\usepackage{booktabs}
\usepackage{caption}
\usepackage{natbib}
\usepackage{placeins}
\usepackage{adjustbox}
\usepackage{setspace}
\usepackage{algorithm}
\usepackage{algpseudocode}
\usepackage{multicol} 
\usepackage{makecell, multirow, tabularx} 
\usepackage{rotating}
\usepackage{float}
\usepackage{ulem}
\usepackage{graphicx}
\usepackage{tabularx}
\usepackage{authblk}
\usepackage{svg}
\usepackage[title]{appendix}

\newtheorem{Theorem}{Theorem}[section]
\newtheorem{Proposition}{Proposition}[section]

\newtheorem{Claim}{Claim}[section]
\newtheorem{Lemma}{Lemma}[section]
\newtheorem{Remark}{Remark}[section]

\newtheorem*{Remarknn}{Remark}

\newcommand{\indep}{\perp \!\!\! \perp}

\newcommand{\ceil}[1]{\left\lceil #1 \right\rceil}

\title{Deep Limit Model-free Prediction in Regression}

\author[1]{Kejin Wu}
\author[2]{Dimitris N. Politis}

\affil[1]{Department of Mathematics, University of California, San Diego }
\affil[2]{Department of Mathematics and Halicio\u{g}lu Data Science Institute, University of California, San Diego}

\date{}

\begin{document}

\maketitle

\begin{abstract}
In this paper, we provide a novel Model-free approach based on Deep Neural Network (DNN) to accomplish point prediction and prediction interval under a general regression setting. Usually, people rely on parametric or non-parametric models to bridge dependent and independent variables ($Y$ and $X$). However, this classical method relies heavily on the correct model specification. Even for the non-parametric approach, some additive form is often assumed. A newly proposed Model-free prediction principle sheds light on a prediction procedure without any model assumption. Previous work regarding this principle has shown better performance than other standard alternatives. Recently, DNN, one of the machine learning methods, has received increasing attention due to its great performance in practice. Guided by the Model-free prediction idea, we attempt to apply a fully connected forward DNN to map $X$ and some appropriate reference random variable $Z$ to $Y$. The targeted DNN is trained by minimizing a specially designed loss function so that the randomness of $Y$ conditional on $X$ is outsourced to $Z$ through the trained DNN. Our method is more stable and accurate compared to other DNN-based counterparts, especially for optimal point predictions. With a specific prediction procedure, our prediction interval can capture the estimation variability so that it can render a better coverage rate for finite sample cases. The superior performance of our method is verified by simulation and empirical studies.

%The targeted DNN is trained by minimizing a specially designed loss function so that the conditional inference of $Y$ through the trained DNN generalizes well even though the trained DNN attempts to interpolate dependent variables. In other words, the randomness of $Y$ conditionally on $X$ can be outsourced to $Z$ through the trained DNN.

\end{abstract}

\section{Introduction}\label{Sec:Intro}
Regression analysis is a fundamental statistical tool. More specifically, it is a process to determine the association between dependent variables $Y\in\mathcal{Y}$ and independent variables $X\in\mathcal{X}$; $\mathcal{Y}\subseteq\mathbb{R}$ and $\mathcal{X}\subseteq\mathbb{R}^d$.  Starting from the simplest univariate linear regression ($d=1$) on describing some biological phenomenon in the 19th century, the regression methods have been developed to analyze practical problems in many areas. Generally speaking, to set up regression, people could assume $X$ and $Y$ have a joint cumulative distribution (CDF) $F$, i.e., 
\begin{equation}\label{Eq:jointDis}
(X,Y) \sim F.    
\end{equation}
Usually, this CDF is unknown, and only observations $\{(X_i,Y_i)\}_{i=1}^n\overset{i.i.d.}{\sim} F$ are available. What participators often do is to estimate a parametric or non-parametric regression model as a bridge to `connect' $X$ and $Y$. To measure the inherent variability in the estimation procedure, people rely on the standard error of estimation or –even better–  the corresponding Confidence Interval (CI). In many real-world applications, people are more concerned about the prediction of $Y_f$ \textit{conditional} on some new independent variables $X_f$. To elaborate, we take the simple linear regression as an example. Suppose the underlying data-generating model (DGM) is:
\begin{equation}\label{Eq:linearReg}
    Y_i = \beta^{\prime}X_i + \epsilon_i;
\end{equation}
$\beta\in \mathbb{R}^d$ is the parameter vector; $\epsilon_i$ are $i.i.d.$ errors with mean 0 and finite variance $\sigma^2$;  We may think $X_i$ are fixed, i.e., fixed design regression, or that they are random but independent of errors, i.e., random design regression; for the latter case, the following analysis shall be understood in terms of conditional probability given $X_i = x_i$ for $i = 1,\ldots,n$.

Under \cref{Eq:linearReg}, the popular $L_2$ optimal prediction of $Y_f$ conditional on $X_f = x_f$ is $\beta^{\prime}x_f$. Unfortunately, $\beta$ is usually unknown. The common strategy is replacing $\beta$ with a consistent estimator $\hat{\beta}$. Asymptotically, $\hat{\beta}^{\prime}x_f$ is the optimal prediction in $L_2$ loss. However, a single-point prediction is left much to be desired in practice. As the role of CI in estimation inference, the Prediction Interval (PI) could measure the forecasting accuracy. Under further assumption on errors, i.e., $\epsilon_i$ are $i.i.d.~N(0,\sigma^2)$, it is well known that the PI below
\begin{equation}\label{Eq:PIlinearregNormal}
\mathcal{I}_{\alpha}(x_f): = \left[\hat{\beta}^{\prime} x_f+t_{n-d, \alpha / 2} \hat{\sigma} \sqrt{1+x_f^{\prime}\left(\bm{X}^{\prime} \bm{X}\right)^{-1} x_f}, \hat{\beta}^{\prime} x_f-t_{n-d, \alpha / 2} \widehat{\sigma} \sqrt{1+x_f^{\prime}\left(\bm{X}^{\prime} \bm{X}\right)^{-1} x_f}\right],
\end{equation}
which has a conditional coverage probability $1-\alpha$, i.e., 
\begin{equation*}
     \mathbb{P}\left(Y_f\in \mathcal{I}_{\alpha}(x_f) \right) = 1 - \alpha,~\text{conditional on~} x_f;
\end{equation*}
here $t_{n-d, \alpha / 2}$ is the lower $\alpha/2$ quantile of the $t$ distribution with $n  - d$ degrees of freedom; $\bm{X}$ is the $n\times d$ design matrix having $X^{\prime}_{i}$ as its $i$-th row; $\hat{\sigma}$ is the usual estimator of $\sigma$. However, if the distribution of error is not normal, the coverage of PI (\ref{Eq:PIlinearregNormal}) will not be $1-\alpha$, even asymptotically. In this situation, one naive alternative is
\begin{equation}\label{Eq:PIlinearreg}
    \widehat{\mathcal{I}}_{\alpha}(x_f):= \left[ \hat{\beta}^{\prime} x_f + \widehat{F}_{\epsilon}(\alpha/2),  \hat{\beta}^{\prime} x_f - \widehat{F}_{\epsilon}(\alpha/2)   \right],
\end{equation}
which has a coverage probability $1-\alpha$ asymptotically, i.e.,
\begin{equation*}
     \mathbb{P}\left(Y_f\in \widehat{\mathcal{I}}_{\alpha}(x_f) \right) = 1 - \alpha,~\text{conditional on}~x_f~\text{as}~n\to\infty;
\end{equation*}
$\widehat{F}_{\epsilon}$ is the empirical CDF of regression residuals; $\widehat{F}_{\epsilon}(\alpha/2)$ is the estimated $\alpha/2$ quantile value of $\epsilon$. Hereafter, we define asymptotically valid PI as any intervals whose coverage probability is equal to the nominal confidence level asymptotically as $n$ tends to infinity.

Nevertheless, the usefulness of PI (\ref{Eq:PIlinearreg}) is limited in the finite sample cases. The reason is two-fold: (a) it completely ignores the estimation variability of $\hat{\beta}$. In other words, PI (\ref{Eq:PIlinearreg}) implicitly assumes that $Var(\hat{\beta}^{\prime} x_f) = \sigma^2\cdot x_f^{\prime}\left(\bm{X}^{\prime} \bm{X}\right)^{-1} x_f$ is negligible which is not the case in finite samples, so PI (\ref{Eq:PIlinearreg}) will undercover $Y_f$ if the sample size is small; (b) all these analyses are based on the assumption that \cref{Eq:linearReg} is the correct DGM. To alleviate the limitation of part (b), people may consider the non-parametric regression, e.g., assuming the DGM is $Y_{i} = \mu(X_i) + \epsilon_i$; $\mu(\cdot)$ is unknown but assumed smooth. Then, the regression analysis can be deployed with the help of the kernel smoothing technique. Despite the non-parametric treatment bringing more flexibility, it still constitutes an additive model form and is thus also restrictive. 
%For example, there is no classical way to capture the estimation/prediction inference of an ambiguous DGM: $Y_i = G(X_i,\epsilon_i)$; $G(\cdot, \cdot)$ is a non-linear/non-additive function of two variables. 
More ambitious, we may ask how to make estimation/prediction inferences without any model assumptions and cover the issue (a) meanwhile.

Fortunately, the Model-free (MF) prediction principle provides a way to shun any model conditions. This MF prediction idea was initially implied in the work \cite{politis2003normalizing} on forecasting financial time series with a so-called Normalizing and Variance Stabilizing method, and then was well explained in \cite{politis2015model}. Subsequently, the detailed estimation and prediction inferences based on this MF approach were developed by \cite{wang2021asymptotic} and \cite{wang2021model}, respectively. In short, this MF method transforms the dependent variables $Y_1,\ldots,Y_n$ to a sample of $n$ $i.i.d.$ variables through some kernel estimator. Then, the statistical inference of original dependent variables can be acquired by transforming the corresponding inference of $i.i.d.$ variables inversely; see more details of this idea in \cref{Sec:MFPM} and some related real-world applications in \cite{chen2019optimal, wu2021model, wu2023model}. Unsurprisingly, the current MF prediction method suffers from the curse of dimensionality issues due to the essentials of various kernel estimators, i.e., the convergence rate of the non-parametric estimator gets slower as $d$ gets larger.
%i.e., the optimal rate of convergence for the non-parametric estimation of a smooth regression function is $n^{-2p/(2p+d)}$; $p$ measures the smoothness of the underlying regression function; for a fixed $p$, the convergence rate gets slower as $d$ gets larger. 
To overcome this difficulty, many researchers have proposed non- and semi-parametric models with certain structural constraints to reduce the model dimensionality; see the analysis of generative additive regression model in \cite{stone1985additive} and single-index model in \cite{ichimura1993semiparametric} for examples. As we have illustrated, these methods are limited because of the strong model structure assumption.

Recently, Deep Neural Networks (DNN), one of the Machine Learning methods, has attracted increasing attention. Interestingly, it seems that the DNN can conquer the curse of dimensionality as they have great performance for some high-dimensional tasks, e.g., image processing. Researchers were explaining the reason actively. Similar to the classical non-parametric estimation scenario, \cite{bauer2019deep,kohler2019estimation} showed that the convergence rates of the DNN estimator can be improved significantly if the regression model is assumed to have some specific structure and smoothness properties.
%be multivariate adaptive regression splines or in a generalized hierarchical interaction form. 
\cite{nakada2020adaptive} showed that DNN estimators can circumvent the curse of dimensionality by adapting to the intrinsic low dimensionality of the data with the Minkowski dimension. Moreover, they argued that this advantage of DNN is prevailing broader compared to standard kernel methods that are also adaptive to an intrinsic dimension; see more related discussions in \cref{Subsec:errorboundDNNunicontinuous}. Without other specifications, the term DNN in this work refers to the standard fully connected feedforward DNN with ReLU activation functions.

In this paper, we replace the kernel estimators in the standard MF prediction method with DNN estimators. This replacement will kill two birds with one stone: (1) According to folk wisdom and empirical/theoretical evidence, DNN estimators suffer less from the curse of dimensionality. Thus, the MF prediction with the help of DNN shall be more accurate when we have multivariate independent variables and the sample size is small; (2) DNN, as an unstable estimator may vary widely across different samples, so its stability should be considered. The MF prediction principle provides a way to capture the estimation variability of DNN when the predictions are required. As a result, we develop a so-called \textit{Deep Limit Model-free} (DLMF) prediction method. It is a unified prediction framework that can cooperate with various DNN models on different application scenarios. In the context of this paper, we focus on the prediction inference of standard regression problems and attempt to construct a better PI in the finite sample cases. To distinguish our PI from other ``naive'' PI, e.g., PI (\ref{Eq:PIlinearreg}), we will show our PI has \textit{pertinent} property, i.e., the ability to capture the estimation variability; the definition will be given later. 

We should also notice that there are other Machine Learning methods to explore conditional prediction inference under setting (\ref{Eq:jointDis}), e.g., conditional Generative Adversarial Networks (cGAN) proposed by \cite{mirza2014conditional}. cGAN is usually conditional on the class label information and then attempts to learn the distribution of data. However, as revealed by \cite{zhou2023deep}, cGAN does not work well for data generation with continuous conditional variables. Inspired by the adversarial training procedure, they proposed an alternative DNN-based distribution learner and called it \textit{Deep Generator}; \cite{liu2021wasserstein} made further improvements to the Deep Generator by applying Wasserstein distance in the training process. Our DLMF prediction implies a simpler training algorithm and is more stable than various Deep Generators. More importantly, our DLMF prediction framework furnishes us with a way to capture estimation variabilities in predictions.

The paper is organized as follows. In \cref{Sec:MFPM}, we present the MF prediction principle and show its general feasibility. Next, we develop the Deep Limit Model-free prediction method in \cref{Sec:DLMFP}. In addition, a new high probability error bound for uniformly continuous function is given there. %Some practical issues in performing DLMF prediction are discussed in \cref{Sec:Tuningpara}. 
The pertinent PI based on the DLMF prediction framework is explored in \cref{Sec:PPI}. Simulation and empirical studies are given in \cref{Sec:Simulation} and \cref{Sec:empiricalcase}, respectively. We summarize this paper in \cref{Sec:Conclu}. Complementary materials and proofs are given in \hyperref[Appendix:A]{Appendix A} and \hyperref[Appendix:B]{Appendix B}, respectively.

\section{Model-free prediction method}\label{Sec:MFPM}
For completeness, we now give a brief introduction to the MF prediction principle; a complete tutorial can be found in \cite{politis2015model}. Then, we will introduce one variant of the MF prediction method--Limit MF prediction. 

\subsection{Model-free prediction principle}
In short, the MF prediction principle relies on an invertible transformation function that maps non-$i.i.d.$ dataset to a dataset consisting of $i.i.d.$ variables. 
%Then, the prediction inference of $Y_f$ can be studied by transforming the prediction inference of $i.i.d.$ variables inversely. 
For convenience, we re-state the regression setup of this paper here:
\begin{equation}\label{Eq:RegsetupSec2}
    n \text{~samples~} \{(X_i,Y_i)\}_{i=1}^n\overset{i.i.d.}{\sim } F.
\end{equation}
We should notice that the above setup implies a random design. As we discussed in \cref{Sec:Intro}, the fixed design can be analyzed similarly in terms of conditional probability given $X_i = x_i$. 

Although the jointly $i.i.d.$ condition implies that $\{X_i\}_{i=1}^n$ and $\{Y_i\}_{i=1}^n$ are all marginally $i.i.d.$, $\{X_i\}_{i=1}^n$ and $\{Y_i\}_{i=1}^n$ are not necessarily mutually independent to each other. People can take various parametric or non-parametric models to model the dependence between $X$ and $Y$. In the spirit of MF prediction principle, we do not assume any latent regression model, but we aim to find invertible transformation functions $H_{X_i}$ which transform each $Y_i$ to $Z_i$ conditional on $X_i$ for $i=1,\ldots,n$; $Z_i$ has a simple cumulative distribution $F_Z$ conditional on $X_i$, e.g., Uniform[0,1] or standard normal. We use the subscript of $H_{X_i}$ to indicate that the transformation function depends on $X_i$ implicitly. In other words, we can think that the collection of functions $\{H_{X_i}\}_{i=1}^n$ connects two equivalent probability spaces: one is the space of $\{(X_i,Y_i)\}_{i=1}^n$ and the other is the space of $\{(X_i,Z_i)\}_{i=1}^n$; here $X_i$ and $Z_i$ are mutually independent. The reason for this claim is that each $Z_i$ has the same distribution $F_{Z}$, e.g., Uniform[0,1], conditional on $X_i$ for all $i$, so $Z_i$ are independent with $X_i$. It further has the unconditional distribution $F_Z$.

We may now assume that desired $\{H_{X_i}\}_{i=1}^n$ exist. Our ultimate goal is the prediction inference of $Y_f$ conditional on $X_f = x_f$ given the regression setup (\ref{Eq:jointDis}) solely. Denote the distribution of $Y_f$ conditional on $x_f$ by $F_{Y|x_f}$. Under the Model-free prediction idea, this conditional distribution can be recovered by transforming the distribution of $Z_f$ conditioning on $x_f$. The key observation is that the random variable $H_{x_f}^{-1}(Z_f)$ shall have the same cumulative distribution as $F_{Y|x_f}$; here $Z_f\sim F_Z$. Again, the subscript of the inverse transformation function $H_{x_f}^{-1}$ indicate the implicit dependency between $x_f$ and $H$. If $H_{x_f}^{-1}$ and $F_Z$ are known, we may determine the analytical expression $H_{x_f}^{-1}(Z_f)$, and then we can derive the prediction inference of $Y_f$ exactly. However, it may be intractable to obtain such an analytical form. In this situation, the Monte Carlo method can provide an alternative way to get the prediction inference of $Y_f$. More specifically, we can sample $\{Z_j^*\}_{j = 1}^{M} \overset{i.i.d.}{\sim} F_Z$; $M$ is a  large enough constant. $F_{Y|x_f}$ can be approximated by the empirical distribution $F_{\text{S}}$ which is based on $ \{H_{x_f}^{-1}(Z_j^*)  \}_{j=1}^M$. By the Glivenko–Cantelli theorem, it is easy to find that $F_{\text{S}}$ converges to $F_{Y|x_f}$ uniformly when $M\to \infty$\footnote{In practice, $H_{x_f}$, $H_{x_f}^{-1}$ and $F_Z$ may be unknown. We have to find appropriate estimations $\widehat{H}_{x_f}$, $\widehat{H}_{x_f}^{-1}$ and $\widehat{F}_Z$. Then, the simulation-based method needs to be replaced with a bootstrap-based method. We can sample $\{Z_j^*\}_{j=1}^{M} \overset{i.i.d.}{\sim} \widehat{F}_Z$; $F_{Y|x_f}$ can be approximated by the empirical distribution $F_{\text{B}}$ which is the limiting distribution of $ \{\widehat{H}_{x_f}^{-1}(Z_j^*)\}_{j=1}^M$ when $M$ tends to infinity. As $n\to \infty$, we can verify that $F_{\text{B}}$ converges to $F_{Y|x_f}$ uniformly under mild conditions.}.

Once $F_{Y|x_f}$ is determined exactly based on the analytical expression $H_{x_f}^{-1}(Z_f)$ or approximated by $\widetilde{F}_{Y|x_f}:= F_S$, the optimal $L_2$ and $L_1$ conditional prediction of $Y_f$ are the mean of $F_{Y|x_f}$ or $\widetilde{F}_{Y|x_f}$ and the median of $F_{Y|x_f}$ or $\widetilde{F}_{Y|x_f}$, respectively. For the equal-tail PI $\mathcal{I}_{\alpha}(x_f) :=  [F^{-1}_{Y|x_f}(\alpha/2), F^{-1}_{Y|x_f}(1-\alpha/2)]$, it covers $Y_f$ with exact $1-\alpha$ confidence level, i.e.,
\begin{equation}\label{Eq:oraclePI}
    \mathbb{P}(Y_f \in \mathcal{I}_{\alpha}(x_f) ) = 1-\alpha.
\end{equation}
For the equal-tail quantile PI $\widetilde{\mathcal{I}}_{\alpha}(x_f) :=  [\widetilde{F}^{-1}_{Y|x_f}(\alpha/2), \widetilde{F}^{-1}_{Y|x_f}(1-\alpha/2)]$, it is asymptotically valid with $1-\alpha$ nominal confidence level, i.e.,
\begin{equation}\label{Eq:pertinentPI}
    \mathbb{P}(Y_f \in \widetilde{\mathcal{I}}_{\alpha}(x_f) ) = 1-\alpha,~\text{asymptotically conditional on}~x_f~\text{as}~M\to\infty.
\end{equation}

\begin{Remark}\label{Remark:relaPI}
    At first glance, PI in \cref{Eq:oraclePI} is nothing else than the quantile PI based on the known conditional CDF $F_{Y|x_f}$. However, our procedure to get this PI is intrinsically different from taking naive quantile PI. We rely on an MF transformation approach. This is the key component that empowers our PI to capture the estimation variability if it exists in practice. More discussions are given in subsequent sections. 
\end{Remark}

Recall that PI (\ref{Eq:PIlinearreg}) in \cref{Sec:Intro} is also asymptotically valid. However, there should be a difference between PI (\ref{Eq:PIlinearreg}) and PI in \cref{Eq:pertinentPI} for finite sample cases. The reason is that there is unaccounted estimation variability within PI (\ref{Eq:PIlinearreg}). On the other hand, there is no estimation issue in PI of \cref{Eq:pertinentPI} since we assume that inverse transformation function $H_{x_f}^{-1}$ is known.

In short, we will call PI asymptotically pertinent if there is no estimation variability (e.g., PI in \cref{Eq:pertinentPI} or the variability can be captured in some way; see the formal definition of pertinent PI in \cref{Sec:PPI}. Later, we will see that the oracle PI in \cref{Eq:oraclePI} is hardly achievable, and pertinent PI is the best candidate that we could pursue. To conclude this section, we divide the MF prediction procedure into two stages and summarize them in \cref{Table:schematic}. A detailed MF prediction algorithm will be presented in \cref{Subsec:DLMFM}. 
\begin{table}[htbp]
    \centering
    \begin{tabular}{ccc}
    \toprule
     Transformation Stage   & \hspace{30pt} &  Prediction Stage\\
     $Y_i \xrightarrow{H_{X_i}} Z_i  \overset{i.i.d.}{\sim} F_Z$ &   &  $Z_f\sim F_Z  \xrightarrow{H^{-1}_{x_f}} Y_f  \sim  F_{Y|x_f}$\\
     \bottomrule
    \end{tabular}
    \caption{A schematic of the MF prediction method.}
    \label{Table:schematic}
\end{table}

\subsection{The existence of transformation function}\label{Subsec:ExistenceofH}
So far, we assume that we know transformation functions $\{H_{X_i}\}_{i=1}^n$. It is still unclear if this function collection exists. Fortunately, the Probability Integral Transform (PIT) provides the possibility to determine $\{H_{X_i}\}_{i=1}^n$; it turns out that they could be conditional CDFs $\{F_{Y|X_i} \}_{i=1}^n$ under regularity conditions; see \cite{angus1994probability} for related discussions about the PIT. Interestingly, the conditional CDF $F_{Y|X}$ is exactly our goal to make prediction inferences. As we have mentioned in \cref{Remark:relaPI}, the naive quantile PI and our MF PI coincide under the ideal situation, i.e., known $F_{Y|X}$. However, our MF PI could have the \textit{pertinent} property for the cases in which $F_{Y|X}$ is unknown; see \cite{politis2015model,wang2021model} for more discussions. To elaborate a little further, let's suppose we only have information about $n$ samples $\{(X_i,Y_i)\}_{i=1}^n$ from $F$ as the setup (\ref{Eq:RegsetupSec2}). In this situation, the PI in \cref{Eq:oraclePI} can not be simply obtained even viewing it from the naive quantile PI perspective. It is tempting to replace $F_{Y|x_f}$ in PI with $\widehat{F}_{Y|x_f}$; although this plugged-in type PI is asymptotically valid once $\widehat{F}_{Y|x_f}$ is a consistent estimator to $F_{Y|x_f}$, it does not capture the estimation variability raised from estimating $F_{Y|x_f}$ by $\widehat{F}_{Y|x_f}$ when the sample is finite. As a result, the naive PI suffers from the notorious undercoverage issue. In this section, we discuss the existence of the transformation function. A kernel-based estimator is given. The comparison of naive PI and our MF PI will be given in \cref{Sec:Simulation}. 

To apply PIT, we make the below assumption:

\begin{itemize}
    \item [A1] The marginal density $f_x$ of $X$ and the conditional CDF $F_{Y|X}(y|x)$ have continuous second order derivations with respect to $x$. $F_{Y|X}$ is Lipschitz continuous in $y$ for all $x$; the conditional density $f_{Y|X}(y)$ is non-zero for all $y$ and $x$; $f_x$ is non-zero for all $x$. 
\end{itemize}
Define the kernel estimator of $F_{Y|X}$ based on $\{(X_i,Y_i)\}_{i=1}^n$ below:
\begin{equation}\label{Eq:kernelEst}
    \widehat{F}_{Y|X}(y|x) =\frac{\frac{1}{n} \sum_{i=1}^n W_h\left(X_i, x\right) K\left(\frac{y-Y_i}{h_0}\right)}{\overline{W}_h(x)};
\end{equation}
$K$ is a smooth (continuously differentiable) distribution function that is strictly increasing, e.g., a proper CDF; $W_h\left(X_i, x\right)=\frac{1}{h_1\cdot h_2\cdots h_d} \prod_{s=1}^d w\left(\frac{X_{i,s}-x_s}{h_s}\right)$ and $\overline{W}_h(x)=\frac{1}{n} \sum_{i=1}^n W_h\left(X_i, x\right); w(\cdot)$ is a univariate, symmetric density function with bounded support; $X_{i,s}$ and $x_s$ are the $s$-th coordinate of $X_i$ and $x$; $\{h_i\}_{i=0}^d$ are bandwidths. The kernel estimator $\widehat{F}_{Y|X}(y|x)$ in \cref{Eq:kernelEst} also smoothes the dependent variables. The benefit is that $\widehat{F}_{Y|X}(y|x)$ is also continuous and strictly increasing in $y$. For simplicity, we take $h = h_1 = h_2 = \cdots = h_d$. To make $\widehat{F}_{Y|X}(y|x)$ be consistent to $F_{Y|X}$, we assume:
\begin{itemize}
    \item[A2] $h\to 0$, $nh\to\infty$; $h_0\to 0$; $(nh^d)^{1/2}(h_0^3 + dh^3) \to 0$.
\end{itemize}
By Theorem 6.2 of \cite{li2007nonparametric}, we can get the lemma below:
\begin{Lemma}\label{Lemma:KernelEst}
    Under A1 and A2, $\widehat{F}_{Y|X} \xrightarrow{p} F_{Y|X}$, More specifically, $Var(\widehat{F}_{Y|X}) = O(\frac{1}{nh^d})$ and $Bias(\widehat{F}_{Y|X}) = O(dh^2)$.
\end{Lemma}

Treating $\widehat{F}_{Y|X}$ as the transformation function $H_{x_f}$, we define the transformed variables $\Gamma_i := \widehat{F}_{Y|X}(Y_i|X_i)$ for $i=1,\ldots,n$. On the other hand, we define the oracle variables $Z_i := F_{Y|X}(Y_i|X_i)$. To simplify the proof and meet the later optimization requirement related to the DNN estimator, i.e., we require that the metric entropy of DNN space be a bounded number, we assume:

\begin{itemize}
    \item[A3] The domain of $Y$ and $X$ are compact sets, respectively, i.e., $\mathcal{Y} := [-M_1,M_1]$ and $\mathcal{X} := [-M_2,M_2]^d$; $M_1$ and $M_2$ are two constants.
\end{itemize}
\begin{Remark}  \label{Remark:truncated}
Assumption A3 gives us the convenience of applying the existing convergence result of DNN estimators directly. We could apply a weaker assumption such that $P\left(|Y|>\tau\right) \leq C \rho_1^{-\tau}$ and $P\left(\left\|X\right\|>\tau\right) \leq C \rho_2^{-\tau}$ for some appropriate $\rho_1$ and $\rho_2$ (sub-exponential). Then, the event that $X$ and $Y$ belong to a compact set has a \textit{high probability}. 
\end{Remark}

In consequence, $\{\Gamma_i\}_{i=1}^n$ are consistent to $\{Z_i\}_{i=1}^n$ for each $i$ and $X_i$. We summarize this fact as:
\begin{Lemma}\label{Lemma:consistencyEtaZ}
    Under A1-A3, $\Gamma_i \xrightarrow{p} Z_i$ for each $i$, where $Z_i$ is an $\textit{Uniform}(0,1)$ random variable.
\end{Lemma}
In short, the proof of \cref{Lemma:consistencyEtaZ} depends on the uniform consistency of $\widehat{F}_{Y|X}(y|x)$, i.e., $\sup_y|  \widehat{F}_{Y|X}(y|x) - F_{Y|X}(y|x) | \xrightarrow{p} 0 $ for all $x$. This is trivial under A1-A3, the smoothness assumptions of $K(\cdot)$ and $F_{Y|X}$, and the pointwise consistent of $\widehat{F}_{Y|X}(y|x)$ to $F_{Y|X}(y|x)$. As a result, $\Gamma_i \xrightarrow{p} Z_i$ for each $i$.

\cref{Lemma:consistencyEtaZ} implies that PI in \cref{Eq:pertinentPI} is still asymptotically valid after replacing $\{Z_i\}_{i=1}^n$ by $\{\Gamma_i\}_{i=1}^n$. If we further assume that $F_{Y|X}$ is strictly increasing at point $y = F^{-1}_{Y|X}(\alpha)$ for some $\alpha$ and all $x$, then we can approximate $F^{-1}_{Y|x_f}(\alpha)$ by $\widehat{F}^{-1}_{Y|x_f}$. As a result, PI in \cref{Eq:pertinentPI} is still asymptotically valid even with an unknown transformation function. However, we need further steps to capture the estimation variability so that the asymptotically pertinent property can be recovered. More details are given in \cref{Sec:PPI}.  

\section{Deep Limit Model-free prediction}\label{Sec:DLMFP}
As discussed in the previous section, the MF prediction method is based on an invertible transformation function. Under the ideal situation ($F_{Y|X}$ is known), the PI of MF prediction is equivalent to the simple quantile PI. However, a more common situation is that $F_{Y|X}$ must be estimated before making a prediction inference. \cref{Subsec:ExistenceofH} shows the feasibility of applying MF prediction with kernel estimators. To utilize the flexibility of DNN, we attempt to reproduce the MF prediction method with DNN estimators and then develop the so-called \textit{Deep Limit Model-Free} (DLMF) method; the motivation and corresponding algorithm are given in \cref{Subsec:DLMFM}; the approximation ability to apply DNN to estimate a (uniformly) continuous function is analyzed in \cref{Subsec:errorboundDNNunicontinuous}; other DNN-based counterparts and the comparisons with our methods are described in \cref{Subsec:otherDNNcounter,Subsec:preliminaryanalysis}; the formal theoretical analysis of DLMF method is given in \cref{Subsec:AsyanalysisDNN}.

\subsection{Deep Limit Model-free method}\label{Subsec:DLMFM}
DNN was shown to possess a strong approximation power. For example, the fully connected feedforward DNN is a universal estimator for any function $f$ in an H\"{o}lder space, i.e., $ || f - f_{\text{DNN}} ||_{\infty} \to 0$ once the size (which is defined as the total number of parameters within DNN) of a DNN estimator $f_{\text{DNN}}$ converges to infinity; see \cite{yarotsky2018optimal, yarotsky2020phase} for discussions. Unsurprisingly, DNN could be a good estimator for smooth $F_{Y|X}$, but it may be intractable to find the inverse of the standard feedforward DNN. In addition, another critical issue is how can we guarantee the $i.i.d.$ property of transformed variables $\{\Gamma_i\}_{i=1}^n$ conditional on each $X_i$. The kernel estimator satisfies this requirement due to PIT. On the other hand, DNN is a ``black box'' and shall map $\{(X_i,Y_i)\}$ together to $i.i.d.$ variables to satisfy the conditional transformation requirement. To get the desirable DNN estimator, we usually need to optimize the objective functions with a specific algorithm, e.g., Stochastic Gradient Descent. In the context of this paper, the objective function can be some ``distance'' measure to evaluate the difference between transformation results and $i.i.d.$ $\{Z_i\}_{i=1}^n$. However, $i.i.d.$ property of $\{\Gamma_i\}_{i=1}^n$ by DNN is hard to check. 
%One possible way is transforming $\{(X_i,Y_i)\}$ to Gaussian first as a ``stepping stone''. Then, the whitening manipulations shall give $i.i.d.$ transformed variables finally. Undoubtedly, this two-step transformation brings unnecessary difficulty. 

Before starting our methods, we should notice that there are some specific machine learning methods to achieve a similar distribution transformation purpose. For example, the flow-based methods consist of a sequence of specifically designed transformation layers so that the whole transformation process can be invertible. The components of each layer are usually the outcomes of a DNN; see \cite{huang2018neural,winkler2019learning} for references. In addition, variational autoencoder (VAE) and its variant conditional VAE are tools to recognize samples by an encoder and then generate new samples via a decoder. Usually, the encoder and decoder are normal distributions with mean and variance estimated by DNN; see \cite{kingma2013auto, pagnoni2018conditional} for references. However, these two approaches have their limitations due to specific flow structures and encoder/decoder distributions. Our transformation method is not spoiled by these drawbacks. 

To explain our method, the key observation is that the forward transformation step in the procedure of the MF prediction method is not necessary. In other words, we can do sampling based on $F_Z$ to derive the inference of our target $Y_f$ by $H_X^{-1}$. This approach is the so-called Limit Model-free (LMF) prediction of \cite{politis2015model} because the exact limiting distribution of $\{Z_i\}_{i=1}^n$---$F_Z$ is applied.
% \begin{Remarknn}
%     Although the LMF does not require the knowledge of transformation function $H_X$, it tends to work slightly worse than a Predictive Model-free (PMF) prediction method according to the coverage rate of PI. In short, PMF approach defines transformed variables as $\widehat{U}_t^{(-j)}=\widehat{F}_{Y|X}^{(-j)}\left(Y_j | X_j\right)$ where $\widehat{F}_{Y|X}^{(-j)}$ is estimated through the delete-$j$ dataset $\left\{\left(X_i, Y_i\right)\right\}_{i=1}^n /\left(X_j, Y_j\right)$; see Chapter 4.5 of \cite{politis2015model} for more details. In this paper, we focus on the LMF with DNN. We leave the development of the Deep PMF prediction method to future work. 
% \end{Remarknn}
The feasibility of LMF prediction is easy to check because PIT always guarantees the existence of $H_X$ under regular conditions. Then, a DNN could approximate $H_X^{-1}$ well assuming it is continuous. Here, we rely on a stronger result to define one inverse transformation function in the Model-free prediction procedure. Denote the targeted inverse transformation function by $H_{0}: \mathcal{X} \times \mathcal{Z} \to \mathcal{Y}$; $\mathcal{Z}$ is the domain of transformed variables; it could be $[0,1]$ for uniform variables or $\mathbb{R}$ for normal variables; we wish that $H_{\text{DNN}}$ takes $X$ and $Z$ as input and returns values in the domain of $\mathcal{Y}$. We may consider multidimensional reference variable $Z$, e.g., $[0,1]^p$ or $N(0,I_p)$. The dimension $p$ could be a tuning parameter to control the empirical coverage rate of resulting PI; see related simulation and empirical studies in \cref{Sec:Simulation} and \cref{Sec:empiricalcase}. 

The theoretical foundation of our DLMF method relies on one variant of the noise-outsourcing lemma which is Lemma 5 of \cite{bloem2020probabilistic}:

\begin{Lemma}[one variant of the noise-outsourcing lemma]\label{Lemma:variantofnoiseout}
    Let $X$ and $Y$ be random variables with joint distribution $P_{X, Y}$. Let $\mathcal{S}$ be a standard Borel space and $S: \mathcal{X} \to \mathcal{S}$ a measurable map. Then $S(X)$ $d$-separates $X$ and $Y$, i.e., $Y \indep_{S(X)} X$ if and only if there is a measurable function $\Xi:\mathcal{S} \times [0,1]   \to \mathcal{Y}$ such that
$$
(X, Y) \stackrel{a.s.}{=}(X, \Xi(S(X), Z)),~ \text { where } Z \sim \text{Uniform}[0,1] \text { and } Z \indep X;
$$
                     
\end{Lemma}
``$\stackrel{a.s.}{=}$'' denotes equal almost surely; $Z \indep X$ represents that $Z$ and $X$ are independent; $Y \indep_{S(X)} X$ means that $Y$ and $X$ are conditional independent given $S(X)$. More specifically this conditional independence is held given the $\sigma$-algebra generated by $S(X)$. In the context of this paper, we can take $\mathcal{X}$ and $\mathcal{Y}$ in \cref{Lemma:variantofnoiseout} as $[-M_2,M_2]^d$ and $[-M_1,M_1]$, respectively, under A3. Moreover, if we take $S(X) = X$, then $Y \stackrel{a.s.}{=} \Xi(Z, X)$ and $Y$ has conditional distribution $F_{Y | X}$. The original version of the noise-outsourcing lemma is Theorem 6.10 of \cite{kallenberg2002foundations} in which the theorem name is \textit{transfer}. The proof of \cref{Lemma:variantofnoiseout} is trivial based on the Proposition 6.13 of \cite{kallenberg2002foundations}. 

Then, we define the optimal inverse transformation function as:
\begin{equation}\label{Eq:empriskDNNTrue}
    H_0 = \arg\min_{H}\mathbb{E}\left( Y- H(X,Z)  \right)^2;
\end{equation}
It turns out that the optimal solution of the above optimization problem is $H_0(X,Z) \stackrel{a.s.}{=} Y$. Similarly, $F_{Y|x_f}$ can be represented by the conditional distribution of $H_0(x_f,Z)$. The existence of oracle $H_0$ is guaranteed by \cref{Lemma:variantofnoiseout} where a satisfied measurable function $\Xi(\cdot,\cdot)$ is provided. However, we hope $\Xi$ can possess some smoothness property, at least $C^0$, to be well estimated by DNN. If we let $A: = [-M_2, M_2]^d\times [0,1]$, we can find a continuous $\widetilde{\Xi}: A \to [-M_1,M_1]$ such that $\widetilde{\Xi}(x,u) = \Xi(x,u)$ for all $(x,u)\in D \subseteq A$; here $\lambda(A\backslash D)<\epsilon$ for $\forall \epsilon>0$. In other words, we can take $H_0(\cdot,\cdot)$ as $\widetilde{\Xi}(\cdot,\cdot)$ after ignoring a negligible set. The replacement of the $\text{Uniform}[0,1]$ random variable $U$ by other appropriate univariate/multivariate reference random variables is possible and does not affect the continuous property of $\widetilde{\Xi}$. We summarize this fact in \cref{Prop:existenceofH}.  

% The direct result of this lemma is that a measurable function $\Xi(\cdot,\cdot)$ exists so that $Y \stackrel{a.s.}{=} \Xi(U,X)$ for $(X,Y)\sim F$ and $X \indep U\sim \text{Uniform}(0,1)$. Then, with Lusin's and Tietze extension theorems, we can get a continuous $\widetilde{\Xi}(\cdot,\cdot): A: = \mathcal{X}\times [0,1] \to \mathcal{Y}$ such that $\widetilde{\Xi}(x,u) = \Xi(x,u)$ for all $(x,u)\in D \subset A$; here $\lambda(A\backslash D)<\epsilon$ for $\forall \epsilon>0$; $\lambda$ denotes the Lebesgue measure. Finally, we can take $H_0(\cdot,\cdot)$  as $\widetilde{\Xi}(\cdot,\cdot)$ up to a negligible set. 

\begin{Proposition}\label{Prop:existenceofH}
Under the regression set up \cref{Eq:jointDis} and A3, there is a continuous $\widetilde{\Xi}(\cdot,\cdot): A: = [-M_2, M_2]^d\times \mathcal{U} \to [-M_1, M_1]$ such that $\widetilde{\Xi}(x,u) = H_0(x,u)$ for all $(x,u)\in D \subseteq A$; here $\lambda(A\backslash D)<\epsilon$ for $\forall \epsilon>0$; $\lambda$ denotes the Lebesgue measure; $\mathcal{U}$ could be $\mathbb{R}^p$ or $[0,1]^p$ if we take $Z$ as $N(0,I_p)$ or $\text{Uniform}[0,1]^p$, respectively, for some positive integer $p$.
\end{Proposition}

% \begin{Remarknn}[{\color{cyan} Something need more work }][Remove this remark for this moment]
%     In the beginning, I thought the conditional quantile function $F_{Y|X}^{-1}$ is one candidate of $H_0$. However, this claim may ask too much from PIT. The PIT says that $F_{Y|X}^{-1}(U)$ should have the same distribution as $Y$ conditional on $X$; $U$ is Uniform(0,1). However, we need $H_0(X,Z) \equiv Y$ to minimize \cref{Eq:empriskDNNTrue}. {\color{cyan} I feel that we may prove that $F_{Y|X}^{-1}$ indeed satisfy the minimization condition of \cref{Eq:empriskDNNTrue}, and then we can give necessary conditions to guarantee the smoothness of $H_0$ by assuming that $F_{Y|X}^{-1}$ and $F_{Y|X}$ have some good properties. For this moment, I present a safe way to argue that the minimizer of \cref{Eq:empriskDNNTrue} exists and could be continuous and then be estimated well by DNN.}  )
% \end{Remarknn}

%Meanwhile we hope that the smoothness of $F_{Y|X}^{-1}$ is the same as $F_{Y|X}$. Thus, we make one more assumption on the conditional distribution of $Y$:
%\begin{itemize}
%    \item[A4] $F_{Y|X}$ is strictly increasing w.r.t. $y$ for all $x$.
%\end{itemize}
%The assumption A4 is slightly stronger than the ``local'' strictly increasing condition at the end of \cref{Subsec:ExistenceofH} which is used to show the feasibility of kernel-based transformation function. With A1, A3 and A4, $F_{Y|X}^{-1}$ is also continuous. 

Hereafter, we try to apply DNN to estimate $\widetilde{\Xi}(\cdot,\cdot)$ which is continuous and \textit{nearly} equal to $H_0$. Therefore, our theoretical results will be held except for a negligible set. We should mention that $H_0$ may not be unique since there may exist another $H^{\prime}_0$ such that \cref{Eq:empriskDNNTrue} is still satisfied in the a.s. sense. In other words, \cref{Eq:empriskDNNTrue} defines an equivalence class of functions. For prediction purposes, what matters is that $F_{Y|x_f}$ can be recovered by $H_0(x_f, Z)$. Thus, the non-uniqueness is not a big issue in our applications.

\begin{Remarknn}
   In the standard MF prediction method, the inverse transformation is applied with kernel-based methods. The transformation function is also not unique. For example, if we already have figured out a kernel estimator $K$ which transforms $\{Y_i\}_{i=1}^n$ to $i.i.d.$ $ \{Z_i\}_{i=1}^n \sim \text{Uniform}[0,1]$. We can consider a transformation function $K^{\prime} = c\cdot K$ which shall transform $\{Y_i\}_{i=1}^n$ to $i.i.d.$ $\text{Uniform}[0,c]$ conditional on $\{X_i\}_{i=1}^n$; $c$ is a positive real number. However, the MF prediction still works since $K^{-1}$ and $(K^{\prime})^{-1}$ transform $\text{Uniform}[0,1]$ and $\text{Uniform}[0,c]$ back to $Y$ conditional on $X$, respectively.  
\end{Remarknn}

Correspondingly, we define the empirically optimal DNN as:
\begin{equation}\label{Eq:empriskDNN}
    \widehat{H} = \arg\min_{ \substack{ H_\theta\in\mathcal{F}_{\text{DNN}} \\ \| H_\theta \|_{\infty}\leq M}}\frac{1}{n}\sum_{i=1}^n\left( Y_i - H_\theta(X_i,Z_i)  \right)^2;
\end{equation}
$\mathcal{F}_{\text{DNN}}$ is functional class containing some DNNs with specific structure; see A4 in \cref{Subsec:AsyanalysisDNN} for more details; $\theta$ denotes the parameter of DNN; $M$ can be arbitrarily large. The condition $\| H_\theta \|_{\infty}\leq M$ is simply a regularization to the optimizer so that the metric entropy of $\mathcal{F}_{\text{DNN}}$ will not diverge. Although this restriction on $H_\theta$ limits its approximation ability, it is not the case in the context of our purpose since A3 implies that the optimal $H_0$ is also bounded or the extreme part is negligible by \cref{Remark:truncated}. Due to the common symmetry properties of DNN, $\widehat{H}$ may not be unique in the space $\mathcal{F}_{\text{DNN}}$. Fortunately, this identifiability issue is not a big problem for prediction, even for PI again; we should mention that the estimation variability captured by PI should be thought of as the estimation error by treating $\widehat{H}$ as a whole term rather than capturing the estimation error for each parameter within DNN. The latter one is impossible without additional assumptions since DNN is not identifiable; see \cite{franke2000bootstrapping} for one bootstrap approach to do estimation inference of the parameters $\theta$ of neural networks once the identification issue is solved after adding some requirement. 

\begin{Remarknn}
    The risk implied in \cref{Eq:empriskDNN} is essentially different from the expected loss for standard regression tasks, e.g., $\mathbb{E}[(Y - h(X))^2]$ which try to approximate the conditional mean of $Y$ given $X$ by $h(X)$. In other words, the minimizer of $\mathbb{E}[(Y - h(X))^2]$ is $h(X) = \mathbb{E}(Y|X)$. This result has a geometric interpretation that $\mathbb{E}(Y|X)$ is the projection of $Y$ onto a closed subspace of $L_2$ consisting of all random variables which can be written in a function of $X$. Thus, we may interpret that the risk in \cref{Eq:empriskDNNTrue} is dedicated to doing a projection of $Y$ onto a space extended by reference random variable $Z$. The conclusions related to the noise outsourcing theorem imply an appropriate extension space exists such that we can find $H_0(X,Z) \stackrel{a.s.}{=} Y$.

    %given a model: $Y = f(X) + \epsilon$; $\mathbb{E}(Y|X) = f(X)$. 
\end{Remarknn}

In practice, we only observe $n$ data pairs $\{(X_i, Y_i)\}_{i=1}^n$. The   $\{Z_i\}_{i=1}^n$ are ``latent'' variables. At first glance, it is unclear how we should set up the empirical risk $\mathcal{R}_n = \frac{1}{n}\sum_{i=1}^n\left( Y_i - H_\theta(X_i,Z_i) \right)^2$ in \cref{Eq:empriskDNN} for our DLMF method. Denote $F_Z$ the common distribution of the reference random variables $Z_i$; due to the presence and non-uniqueness of the oracle transformation $ H_0$, the shape of the distribution  $F_Z$   is not uniquely identifiable.
In other words, $F_Z$   could be taken as $\text{Uniform}[0,1]$ or $N(0,I_p)$ or some other distribution --- with corresponding changes in the shape of  $ H_0$.
Hence, the practitioner may freely choose the shape of the distribution $F_Z$, and subsequently simply sample $i.i.d.~\{Z^{*}_i\}_{i=1}^n$ from the chosen distribution $F_Z$ to create the desired data pairs $\{(X_i, Y_i,Z^{*}_i)\}_{i=1}^n$.
Finally, define the empirical risk $\mathcal{R}_n^* := \frac{1}{n}\sum_{i=1}^n\left( Y_i - H_\theta(X_i,Z^*_i) \right)^2$, if we further denote the minimization of $\mathcal{R}_n^*$ is achieved by $\widehat{H}^*$, then $\widehat{H}^*$ converges to a specific $H_0$ depending on the choice of $F_Z$ in the mean square sense ---  as does $\widehat{H}$. The intuition is that $\mathcal{R}_n^*$ and $\mathcal{R}_n$ are uniformly ``close'' to the true risk $\mathcal{R}:= \mathbb{E}\left( Y- H_0(X,Z) \right)^2$ as $n$ converges to infinity. Then, their argmin's $\widehat{H}^*$ and $\widehat{H}$ should also be ``close'' to $H_0$ as $n\to\infty$. To simplify the notation, we will use $\mathcal{R}_n$ in what follows, but we should keep in mind that $\{Z_i\}_{i=1}^n$ could be directly sampled from $F_Z$.
 
\begin{Remarknn}
We provide a toy example to further motivate our training procedure to get $\widehat{H}$. Suppose we need to estimate the coefficient $\beta$ of a linear regression model $Y = X^{T}\cdot\beta + \epsilon$ with a fixed design based on samples $\{(X_i,Y_i)\}_{i=1}^n$; here, $\epsilon$ has zero mean and finite variance. The ordinary Least Squares (LS)  estimator is  $\widehat{\beta} : =\arg\min_{\beta}\frac{1}{n}\sum_{i=1}^{n}(Y_i - X^{T}_i\cdot\beta)^2$ which is consistent under standard conditions. However, we could also consider the estimator $ \widehat{\beta}^* : = \arg\min_{\beta}\frac{1}{n}\sum_{i=1}^{n}(Y_i - (X^{T}_i\cdot\beta + \epsilon_i^* ))^2$ where $\{\epsilon_i^*\}_{i=1}^n$ are independent of $X$ and can be generated from any distribution with mean zero and finite variance. It is not hard to show that $\widehat{\beta}^*$ is also consistent although the $\widehat{\beta} $ would generally be more efficient.  
Analogously, our DNN-based estimation $\widehat{H}^*$ converges to $H_0$ in the mean square sense even using the artificially generated $\{Z_i^*\}_{i=1}^n$.
\end{Remarknn}

We summarize the DLMF prediction algorithm to do conditional prediction inference, especially for the $L_2$ optimal point prediction in \cref{Algo:DLMFpoint}. More details about the effects of hyperparameters in the algorithm will be given in \cref{Sec:Simulation}.

\begin{algorithm}
\caption{DLMF prediction algorithm for conditional optimal point prediction}\label{Algo:DLMFpoint}
\setstretch{1.25}
\begin{algorithmic}[1]
\Require (a) the training samples $\{X_i,Y_i, Z_i\}_{i=1}^n$, $\{Z_i\}_{i=1}^n\sim F_Z$; (b) the new observed $x_f$; (c) $F_Z$, the distribution of the reference variable $Z$; (d) the learning rate $lr$; (e) the number of epochs; (f) the clipping parameter $m$; (g) the number of evaluations $S$ for the Monte Carlo estimation; (h) the specific optimizer, e.g., SGD, Adam and RMSProp.
\Statex

\State Initiate a DNN $H_\theta\in \mathcal{F}_{\text{DNN}}$ and simulate $\{Z_i\}_{i=1}^n$ from $F_Z$.
\For{number of epochs} 
\State Update $H_\theta$ by descending its stochastic gradient with the chosen optimizer:
$$
\nabla_{\theta} \left\{ \frac{1}{n}\sum_{i=1}^n\left( Y_i - H_\theta(X_i,Z_i)  \right)^2   \right\}
$$
\State Clip the parameter of $H_\theta$ to $[-m,m]$.
\EndFor
\State \textbf{get} The estimated $\widehat{H}(\cdot, \cdot)$.
\State Simulate $\{Z_j\}_{j=1}^S$ from $F_Z$.
\State $\{\widehat{Y}_j\}_{j=1}^{S}$ $\gets$ $\{\widehat{H}(x_f, Z_j)\}_{j=1}^S$.
\State $\widehat{Y}_{f,L_2} = \frac{1}{S}\sum_{j=1}^S \{\widehat{Y}_j\}_{j=1}^{S}$.
\State \textbf{Return} $\widehat{H}(\cdot, \cdot)$, $\{\widehat{Y}_j\}_{j=1}^{S}$ and $\widehat{Y}_{f,L_2}$. 
\Statex
\Statex \textbf{Comments} With $\{\widehat{Y}_j\}_{j=1}^{S}$, $L_1$ optimal point prediction or other quantile estimation of $F_{Y|X_f}$ can be made. 
\end{algorithmic}
\end{algorithm}

\FloatBarrier

\subsection{The approximation ability of DNN on uniformly continuous functions}\label{Subsec:errorboundDNNunicontinuous}

Recently, the non-asymptotic error bound of the least square DNN estimator on estimating regression functions has been studied by \cite{jiao2023deep, farrell2021deep, nakada2020adaptive, schmidt2020nonparametric,bauer2019deep,schmidt2019deep}. These studies rely on different settings: \cite{jiao2023deep, nakada2020adaptive, schmidt2019deep} made distributional assumptions on the independent variables $X$, e.g., $X$ is supported on a low-dimensional manifold, to alleviate the curse of dimensionality; \cite{schmidt2020nonparametric, bauer2019deep} assumed the underlying regression has some special structure. In our context, it is unclear if the optimal transformation function has a desired simple structure. Also, we do not impose any distributional assumptions on $X$; this extension could be a future work. Without any regression model and distribution requirement, \cite{farrell2021deep} showed the nonasymptotic high probability error bound under some regular assumptions:

\begin{equation}\label{Eq:farrellbound}
\left\|\widehat{f}_{\text {DNN }}-f_*\right\|_{L(X)}^2 \leq C \cdot\left\{n^{-\frac{\beta}{\beta+d}} \log ^8 n+\frac{\log \log n}{n}\right\},~\text{with probability at least}~
1-\exp \left(-n^{\frac{d}{\beta+d}} \log ^8 n\right);
\end{equation}
$n$ is the sample size; $d$ is the dimension of input random variables; $\beta$ is the value to measure the smoothness of the underlying true regression function $f_*$ in a Sobolev space; $\| g(x) \|_{L(X)}^2 : = \mathbb{E}_{X}(g(X)^2)$; $\widehat{f}_{\text {DNN }}$ is DNN-based the minimizer of empirical risk. The error bound in \cref{Eq:farrellbound} is based on DNN approximation ability results from \cite{yarotsky2018optimal}. Recently, \cite{wu2024scalable} further improved the error bound by applying the DNN approximation results for functions belonging to an H\"{o}lder space and a so-called scalable subsampling technique.

Recall that we know there is a measurable function $H_0$ such that $H_0(X,Z) \stackrel{a.s.}{=} Y$ by the noise-outsourcing lemma. Then, after applying Lusin’s and Tietze Extension theorems,  we can find a continuous function $\Xi: [-M_2,M_2]^d \times [0,1]^p \to [-M_1,M_1]$, s.t., $\Xi = H_0$ except for a negligible set. By Heine–Cantor theorem, $\Xi$ is uniformly continuous. Unfortunately, no matter whether the error bound in \cref{Eq:farrellbound} or the refined variant in \cite{wu2024scalable} can not be directly taken in our paper since $\Xi$ need to be at least $\alpha$-H\"{o}lder continuous which is a stronger condition than uniformly continuous and then also continuous. Therefore, it is necessary to develop a new error bound for uniformly continuous functions. Subsequently, based on \cref{Theorem:approxerrorofc0} which is Theorem 4.3 of \cite{shen2021deep} and with additional mild conditions on function $\Xi$ listed below:
\begin{itemize}
    \item B1 Let $\omega_f^E(r)$ be the so-called modulus of continuity of a function $f$ on a subset $E$ belongs to the input space $S$:
$$
\omega_f^E(r):=\sup \left\{\left|f\left(x_1\right)-f\left(x_2\right)\right|; d_S\left(x_1, x_2\right) \leq r, x_1, x_2 \in E\right\},~\text{for any }~ r \geq 0.
$$
$d_S(\cdot,\cdot)$ is a metric define in $S$. We assume $\omega_\Xi^E(r) = \Theta(r)$, which means these two quantities have the same order, where $E\subseteq [-M_2,M_2]^d \times [0,1]^p$;
\end{itemize}
we develop a high-probability nonasymptotic error bound for the estimation of $\widehat{H}$ on $\Xi$. These results are presented as follows:
% To get a nonasymptotic estimation error bound, we assume
% \begin{itemize}
%     \item[C1] The domain of $Y$ and $X$ are compact sets, respectively, i.e., $\mathcal{Y} := [-M_1,M_1]$ and $\mathcal{X} := [-M_2,M_2]^d$; $M_1$ and $M_2$ are two constants.
%     \item[C2] The reference random variable $Z$ is $\text{Uniform}[0,1]^p$. (In practice, we may take $Z$ to be $N(0,I_p)$). 
% \end{itemize}

\begin{Theorem}\label{Theorem:errorboundofC0f}
Under A3 and taking reference random variable $Z$ to be $\text{Uniform}[0,1]^p$, let 
$$\widehat{H}:= \arg\min_{ \substack{ H_\theta\in\mathcal{F}_{\text{DNN}} \\ \| H_\theta \|_{\infty}\leq M_1}}\frac{1}{n}\sum_{i=1}^n\left( Y_i - H_\theta(X_i,Z_i)  \right)^2 = \arg\min_{ \substack{ H_\theta\in\mathcal{F}_{\text{DNN}} \\ \| H_\theta \|_{\infty}\leq M_1}} \mathcal{R}_n;$$
$\mathcal{F}_{\text{DNN}}$ is a class of standard fully connected feedforward DNN functions with width $W$ and depth $L$, respectively:
$$W := 3^{d+p+3} \max \left\{(d+p)\left\lfloor N_1^{1 / (d+p)}\right\rfloor, N_1+1\right\}~;~L: = 12 N_2+14+2(d+p);$$ 
see \cref{Remakr:explanationsofWL} for the meaning of $W$ and $L$ for a DNN in this context. We take $N_1 = \ceil{ \frac{n^{\frac{d+p}{2(\tau+d+p)}}}{\log n}}$ and $N_2 = \ceil{ \log(n) } $; $n$ is the sample size; we require $d+p \geq 2$. Then, for large enough $n$, i.e., $n> \max((2eM_1)^2, \text{Pdim}(\mathcal{F}_{\text{DNN}}))$, we have that $ \left\|\widehat{H} - H_0\right\|_{L^2(X,Z)}^2\to 0$ with probability at least $1-\exp(-\gamma)$, as long as $\gamma = o(n)$. Under the further restriction of the modulus of continuity on $\Xi$, i.e., assuming B1, we have that with probability at least $1-\exp(-n^{\frac{d+p-1}{\tau+d+p}})$:
\begin{equation}
   \left\|\widehat{H} - H_0\right\|_{L^2(X,Z)}^2 \leq 
  C\cdot n^{-\frac{2}{\tau+d+p}} +  o(n^{-\frac{2}{\tau+d+p}});~ \text{for}~ d + p \geq 2; \tau>1; (d+p)\left\lfloor N_1^{1 / (d+p)}\right\rfloor \geq N_1+1;
\end{equation}
When $n$ is large enough, we have that with probability at least $1-\exp(-n^{\frac{d+p}{\tau+d+p}})$:
\begin{equation}
   \left\|\widehat{H} - H_0\right\|_{L^2(X,Z)}^2 \leq C\cdot n^{-\frac{2}{\tau+d+p}} + o(n^{-\frac{2}{\tau+d+p}});~\text{for}~d+p\geq 2; \tau>2;(d+p)\left\lfloor N_1^{1 / (d+p)}\right\rfloor < N_1+1;
\end{equation}
where $C$ is a constant whose value may vary from context.
\end{Theorem}

\begin{Remarknn}[Discussion of \cref{Theorem:errorboundofC0f}]
In \cref{Theorem:errorboundofC0f}, $\tau$ is a positive number involved in the error bound of $\left\|\widehat{H} - H_0\right\|_{L^2(X,Z)}^2$. Taking a smaller $\tau$ will result in a smaller error bound. If $d + p = 2 $ and $(d+p)\left\lfloor N_1^{1 / (d+p)}\right\rfloor \geq N_1+1$, it is interesting to see that the order of error bound can be even close to $n^{-1}$. However, this result may be only held for small $n$ cases. When $n$ is large, $d+p$ shall have a same or larger magnitude as $N_1$ to guarantee $(d+p)\left\lfloor N_1^{1 / (d+p)}\right\rfloor \geq N_1+1$. Then, the condition $n> \max((2eM_1)^2, \text{Pdim}(\mathcal{F}_{\text{DNN}}))$ may not be satisfied for a standard fully connected DNN class. Thus, the case $(d+p)\left\lfloor N_1^{1 / (d+p)}\right\rfloor < N_1+1$ will be more common especially for $n$ being large. 
\end{Remarknn}

\begin{Remark}[The width and depth of the DNN in \cref{Theorem:errorboundofC0f}]\label{Remakr:explanationsofWL}
As mentioned in \cref{Sec:Intro}, we focus on the standard fully connected feedforward DNN with ReLU activation functions here. When we say a DNN possesses width $W$ and depth $L$ in \cref{Theorem:errorboundofC0f}, it means the maximum number of neurons among all hidden layers is no more than $W$ and the number of hidden layers of this DNN is no more than $L$. With appropriate approximation results, all theories in this paper can be extended to a general DNN in which all neurons may not be fully and feedforward connected. 

\end{Remark}

\subsection{Predictions using Deep Generators}\label{Subsec:otherDNNcounter}
There are two DNN-based Deep Generators (DGs) proposed by \cite{zhou2023deep} and \cite{liu2021wasserstein}, respectively. Similar to the inverse transformation function in the DLMF prediction framework, DGs also take $X$ and reference variables $Z$ as inputs and return values to mimic $F_{Y|X}$; henceforward we denote the deep generator by $G(\cdot, \cdot)$. 
%As different from DLMF prediction motivated by PIT, their theoretical foundation relies on the Noise-outsourcing lemma; see \cite{austin2015exchangeable} for one version of this lemma. 
Their theoretical foundation also relies on the noise outsourcing theorem.
With the help of the Monte Carlo method, DGs can generate an arbitrary number of pseudo $Y$ values conditional on $x_f$. Then, $\widehat{Y}_{f,L_2}$ and $F_{Y|x_f}$ can be approximated similarly as the DLMF prediction. The training of DGs depends on the adversarial training strategy which was first proposed by \cite{goodfellow2014generative} to build the popular Generative Adversarial Networks (GAN). More specifically, the loss function is based on KL-divergence or Wasserstein-1 distance between joint distribution $(G(X,Z),X)$ and $(Y,X)$. Thus, we call two DGs DG-KL and DG-WA, respectively. To simplify the notation, we use $\widehat{G}_{\text{KL}}$ and $\widehat{G}_{\text{WA}}$ to represent two DNN-based deep generators trained with samples.

We omit a detailed introduction of these two methods and only present two corresponding minimax problems:
\begin{equation*}
    \begin{split}
        (\widehat{G}_{\text{KL}}, \widehat{D}_{\text{KL}}) &= \arg\min_{ \substack{ G_\rho\in\mathcal{F}_{\text{DNN,G}} \\ \| G_\rho \|_{\infty}\leq B_1} } \arg\max_{ \substack{ D_\phi\in\mathcal{F}_{\text{DNN,D}} \\ \| D_\phi \|_{\infty}\leq B_2} } \frac{1}{n}\sum_{i=1}^n D_\phi(G_\rho(Z_i,X_i),X_i) - \frac{1}{n}\sum_{i=1}^n\exp(D_\phi(Y_i,X_i)); \\
        (\widehat{G}_{\text{WA}}, \widehat{D}_{\text{WA}}) & = \arg\min_{ \substack{ G_\rho\in\mathcal{F}_{\text{DNN,G}} \\ \| G_\rho \|_{\infty}\leq B_3} } \arg\max_{ \substack{ D_\phi\in\mathcal{F}_{\text{DNN,D}} \\ \| D_\phi \|_{\infty}\leq B_4} } \frac{1}{n}\sum_{i=1}^n D_\phi(G_\rho(Z_i,X_i),X_i) - \frac{1}{n}\sum_{i=1}^nD_\phi(Y_i,X_i);
    \end{split}
\end{equation*}
$D_\phi$ is the discriminator/critic trained together with generator adversarially; $\{B_j\}_{j=1}^4$ are four arbitrarily large constants. In the original works about DGs, $\{B_j\}_{j=1}^3$ can be allowed to increase slowly with $n$, e.g., $\{B_j\}_{j=1}^3 = \log(n)$. In this paper, we simplify the condition by assuming A3. It should be noted that $B_4$ should be a constant to satisfy the Lipschitz condition of the critic when the dual form of Wasserstein-1 distance is used; see \cite{arjovsky2017wasserstein} for more discussions. In addition, instead of truncating the critic, \cite{gulrajani2017improved} imposed a penalty term about the gradient of the critic to the loss function determined by the dual form of Wasserstein-1 distance. This strategy was also taken by \cite{liu2021wasserstein}. In this paper, we take the truncation way to restrict the $D_\phi$ to meet the Lipschitz condition exactly.

\subsection{Preliminary analysis of DLMF and DGs-based predictions}\label{Subsec:preliminaryanalysis}
We first analyze our DLMF prediction method and two DGs from different aspects. Empirical comparisons will be given in \cref{Sec:Simulation}. For the computability issue, it is obvious that the DLMF is more time-efficient than two DGs based on adversarial training under the same training setting because adversarial training updates two networks one by one. For the stability of the estimator, although the discriminator/critic plays a regularization role to some extent, the adversarial training is pretty unstable and sensitive to basic training settings, e.g., the choice of optimizer and the DNN structure; see \cite{pang2020bag} for a comprehensive study of implementation issues related to adversarial training. More importantly, as implied by the minimax problem, the convergence of DGs relies on the fact that the discriminator/critic attaches the optimal. In other words, the updating of DGs for each step should be conditional on the discriminator/critic being optimal given current DGs. This is usually omitted and hardly achievable in practice. On the other hand, the DLMF method does not suffer from this training issue since the estimation of the inverse transformation function is defined as a convex optimization problem w.r.t. the whole term $H(X,Z)$. Thus, it is guaranteed that there is a unique estimator in the almost sure sense. 

Beyond the computability and stability issues, we argue that the loss function in DLMF is a more appropriate metric compared to the KL divergence and Wasserstein-1 distance. For the KL-divergence, it is well-known to be asymmetric and possibly infinite. Moreover, it crucially depends on the condition that the density function of data exists. In practice, this condition is violated if the distribution of data is supported by a low-dimensional manifold. Then, the distribution of data is only continuous rather than absolutely continuous. For the Wasserstein-1 distance, it seems to be a better loss function than KL-divergence because its computation does not involve density functions and it is a weaker metric than KL-divergence. However, the dual form of Wasserstein-1 distance imposes additional restrictions on critic functions.

For the loss function \cref{Eq:empriskDNN} in the DLMF prediction procedure, if it equals 0 for all $n\to \infty$, it implies that $\widehat{H}(X_i,Z_i) = Y_i$ for all $i = 1,\ldots, n$. This fact further implies that the empirical distribution $(\widehat{H}(X_i,Z_i), X_i)$ uniformly converges to the joint distribution $F_{Y,X}$ almost surely. We summarize this finding below:

\begin{Lemma}\label{Lemma:lossDLMF}
    Under A1, if there is a sequence of functions $\{\widehat{H}_n\}$ such that $\frac{1}{n}\sum_{i=1}^n\left( Y_i - \widehat{H}_n(X_i,Z_i)  \right)^2 = 0$ as $n\to\infty$. Then, $\sup_{y\in\mathcal{Y}, x\in\mathcal{X}}| \widehat{F}_{Y, X}(y, x) - F_{Y,  X}(y, x) | \to 0$ almost surely; $\widehat{F}_{Y, X}$ is the empirical CDF of $\{(\widehat{H}_n(X_i,Z_i), X_i)\}_{i=1}^n$. People also call $\sup_{y\in\mathcal{Y}, x\in\mathcal{X}}| \widehat{F}_{Y, X}(y, x) - F_{Y,  X}(y, x) |$ the Kolmogorov metric between $\widehat{F}_{Y, X}$ and $F_{Y, X}$.
\end{Lemma}
The proof of \cref{Lemma:lossDLMF} is trivial based on the Glivenko–Cantelli theorem. Therefore, minimizing loss in \cref{Eq:empriskDNN} gives a way to approximate the joint distribution which shares a similar spirit with deep generator approaches. The difference is that our approach is close to minimizing the Kolmogorov metric. On the other hand, KL-divergence and Wasserstein-1 distance are all stronger than the Kolmogorov metric indicated in the work of \cite{gibbs2002choosing}. We summarize the relationship between these three metrics as the below claim:
\begin{Claim}\label{Claim:metrics}
    Given a ``background'' probability space $(\Omega,\mathcal{F},\mathbb{P})$, let $\mathcal{M}$ be the space of all probability measures induced by $\mathbb{P}$ on $(\mathcal{Y}, \mathcal{B}(\mathcal{Y}))$, i.e., $\mathbb{P}\circ f^{-1}$ for any measurable function $f: \Omega \to \mathcal{Y}$. Consider two probability measures $\mu$ and $v$ in $\mathcal{M}$, assuming their density function exists, we have: 
    \begin{equation*}
    \begin{split}
    d_K(\mu,v) &\leq (1 + \sup_xg_v(x))\sqrt{d_W(\mu,v)}; \\
    d_W(\mu,v) &\leq \text{diam}(\mathcal{Y})\sqrt{\frac{ d_{KL}(\mu,v)}{2}};
    \end{split}
    \end{equation*} 
    $d_K(\mu,v)$, $d_W(\mu,v)$ and $d_{KL}(\mu,v)$ are the Kolmogorov metric, Wasserstein metric and KL-divergence between $\mu$ and $v$ or two corresponding distribution functions implied by $\mu$ and $v$, respectively; $g_v$ is the density function implied by $v$; $\text{diam}(\mathcal{Y}):= \sup\{d(x,y),x,y\in\mathcal{Y}\}$ is the diameter of $\mathcal{Y}$ for some appropriate metric $d$. As a consequence, given a sequence of distribution $F_{n}$ on $\mathcal{Y}$ and a targeted distribution $F$, $d_W(F_{n},F) \to 0$ implies $d_K(F_n,F) \to 0$ and $d_{KL}(F_{n},F) \to 0$ implies $d_W(F_n,F) \to 0$. 
\end{Claim}
The implication of \cref{Claim:metrics} is that the Kolmogorov metric is weaker than L-divergence and Wasserstein-1 distance. On the other hand, DGs trained based on stronger metrics may be overfitting. Shown in \cref{Sec:Simulation}, our method is more stable than DGs using the Wasserstein metric and KL-divergence. 

\begin{Remarknn}
    For the deep generators approaches, the loss function is used to measure the estimated and underlying joint distributions. However, we should notice that the joint distribution convergence does not imply conditional distribution convergence. Thus, with the help of Pinsker’s inequality, \cite{zhou2023deep} showed that the joint density based on the deep generator converges to the underlying joint density in the total variation norm. Then, the conditional distribution convergence is easily seen under A1. For our method, we take a specifically designed loss function to approximate the inverse transformation function directly and then we can develop the weak convergence result of conditional distribution. More details are given in the next section.
\end{Remarknn}

\subsection{Asymptotic analysis of DLMF}\label{Subsec:AsyanalysisDNN}
In this section, we give the asymptotic analysis of our DLMF prediction. Our goal is to show the uniform convergence of $\widehat{F}_{\widehat{H}(x_f,Z)}$ to $F_{Y|x_f}$; $x_f$ is any future point belongs to $\mathcal{X}$. $\widehat{F}_{\widehat{H}(x_f,Z)}$ is the empirical distribution of $\{\widehat{H}(x_f,Z_i)\}_{i=1}^S$; $S$ is the Monte Carlo sampling number. To achieve the uniform convergence result, the key component is the consistency of $\widehat{H}$ to $H_0$. In the machine learning area, people usually let the size of the DNN increase with the dimension of input and the sample size to make a DNN ($H_{\theta}$) have enough approximation power. On the other hand, we also need the sample size to be larger than $\text{Pdim}(H_{\theta})$ which is the pseudo-dimension of the DNN to develop consistency. Thus, we make the assumption on $\mathcal{F}_{\text{DNN}}$ below:
\begin{itemize}
    \item[A4] Let $\mathcal{F}_{\text{DNN}}$ be the class of forward fully connected neural networks with equal width $W$ and depth $L$ which are given in \cref{Theorem:errorboundofC0f} and their meanings are discussed in \cref{Remakr:explanationsofWL}. We denote any single DNN belongs to $\mathcal{F}_{\text{DNN}}$ by $H_{\theta}$. We require $\text{Pdim}(\mathcal{F}_{\text{DNN}})<n$ and $\| H_\theta \|_{\infty}\leq M $.
\end{itemize}
where $c \cdot \Upsilon L \log (\Upsilon / L) \leq \operatorname{Pdim}(\mathcal{F}_{\text{DNN}}) \leq C \cdot \Upsilon L \log \Upsilon$ for some constants $c, C>0$, found by \cite{bartlett2019nearly}; $\Upsilon$ is the total number of parameters in a DNN. For example, $\Upsilon$ of a $H_\theta$ which belongs to $\mathcal{F}_{\text{DNN}}$ is $(d+p)\cdot W + (L-1)\cdot(W + W^2) + 1 + W$ if all hidden layers have the same number of neurons.  

\begin{Theorem}\label{Theorem:conditionalDFcon}
Under A1 and A3-A4, with probability at least $1-\exp(-n^{\frac{d+p-1}{\tau+d+p}})$ if $(d+p)\left\lfloor N_1^{1 / (d+p)}\right\rfloor \geq N_1+1$ in \cref{Theorem:errorboundofC0f} or with probability at least $1-\exp(-n^{\frac{d+p}{\tau+d+p}})$ if $(d+p)\left\lfloor N_1^{1 / (d+p)}\right\rfloor < N_1+1$, we have:
\begin{equation*}
\sup_{y} \Big|\widehat{F}_{\widehat{H}(x_f,Z)}(y) -   F_{Y|x_f}(y)\Big| \xrightarrow{p} 0,~\text{as}~n\to \infty, S\to \infty;
\end{equation*}
for any $x_f \in \mathcal{X}$ and $y \in \mathcal{Y}$. 
\end{Theorem}
The direct implication of \cref{Theorem:conditionalDFcon} is that we can get consistent $L_2$ optimal point prediction conditional on $X = x_f$. Moreover, we can build asymptotically valid conditional PI by taking appropriate quantile values of $\widehat{F}_{\widehat{H}(x_f,Z)}$. However, as well explained in \cref{Sec:Intro}, the asymptotically valid PI is limited in practice with finite samples. Thus, we attempt to propose a namely pertinent PI to improve the coverage performance in the next section.

\section{Pertinent prediction interval}\label{Sec:PPI}
As alluded to in the introduction, the estimation variability should be captured for prediction tasks. Otherwise, the PI such as a naive quantile PI based on an estimated conditional will suffer from the undercoverage issue with finite samples. DLMF prediction framework offers an opportunity to mitigate the undercoverage problem via a specifically designed procedure. In short, we need to define a quantity namely predictive root as the milestone to get PI. We then attempt to approximate the distribution of the predictive root. Finally, the PI could be built centered at some optimal point prediction.

\subsection{Definition of asymptotic pertinence}
The asymptotic pertinence was first proposed by \cite{politis2015model} formally for Model-based predictions. Subsequently, \cite{wang2021model} gave the corresponding definition in the Model-free prediction context. We summarize the definition as the four conditions below:
\begin{itemize}
\item[C1] Both predictive root $R_f$ and its variant $R_f^*$ conditional on samples can be decomposed into the following representations:
$$
\begin{aligned}
& R_f=\epsilon_f+e_f; \\
& R_f^*=\epsilon_f^*+e_f^*;
\end{aligned}
$$ 
\item[C2] $\sup _x\left|\mathbb{P}^*\left(\epsilon_f^* \leq x\right)-\mathbb{P}\left(\epsilon_f \leq x\right)\right| \xrightarrow{p} 0$ as $n \to \infty$; $\mathbb{P}^*$ is the probability conditional on $\{X_i,Y_i,Z_i\}_{i=1}^n$.  

\item[C3] There is a diverging sequence of positive numbers $a_n$ such that $a_n e_f$ and $a_n e_f^*$ converge to the same nondegenerate distributions, and $\sup _x\left|\mathbb{P}^*\left(a_n e_f^* \leq x\right)-\mathbb{P}\left(a_n e_f \leq x\right)\right| \xrightarrow{p} 0$ as $n \to \infty$.

\item[C4] $\epsilon_f$ is independent of $e_f$ in the real world; similarly, $\epsilon_f^*$ is independent of $e_f^*$ in the world conditional on training samples. 
\end{itemize}
\begin{Remarknn}
    Later, we will show $\epsilon_f$ could be a non-degenerate variable concerning the distribution of future response $Y_f$, and $e_f$ is model estimation error which converges to 0 in probability; $\epsilon_f^*$ and $e_f^*$ are their analogs conditional on training samples.
\end{Remarknn}
In other words, to verify that a PI is indeed asymptotically pertinent, we need to check that C1-C4 are satisfied. 

More specifically, we can let $R_f$ be the predictive root in the real world. For example, $R_f = Y_f - \widehat{Y}_{f,L_2}$; $Y_f\sim F_{Y|x_f}$ and $\widehat{Y}_{f,L_2}$ is the optimal $L_2$ condition point prediction. If we know $F_{R_f}$ which is the distribution of the predictive root $R_f$, a PI centered at $\widehat{Y}_{f,L_2}$ is easy to be built with the help of the quantiles of $F_{R_f}$. However, in practice, this distribution is an oracle since the future distribution $F_{Y|x_f}$ is unknown. Thus, a compromise is that we use another predictive root to approximate $R_f$. In the spirit of bootstrap estimation, we propose $R_f^*$ which is the predictive root conditional on training samples $\{X_i,Y_i, Z_i\}_{i=1}^n$. Analogously to $R_f$, $R_f^* = Y^*_f - \widehat{Y}^*_{f,L_2}$; $Y^*_f = \widehat{H}(x_f,Z^*)$; $Z^* \sim F_Z$; $\widehat{Y}^*_{f,L_2}$ is the optimal $L_2$ point prediction pretending $\widehat{H}$ is the ``true'' inverse transformation function. Since $\widehat{H}$ is known conditional on $\{X_i,Y_i, Z_i\}_{i=1}^n$, we can estimate $F_{R_f^*}$ which is the distribution of $R_f^*$ by a method which is analogous to the estimation in the bootstrap world. Under standard conditions, $F_{R_f^*}$ is a great approximation of $F_{R_f}$. Thus, an asymptotically pertinent PI with $1-\alpha$ coverage rate centered at $\widehat{Y}_{f,L_2}$ could be
\begin{equation}\label{Eq:PPItoyexample}
    \left[\widehat{Y}_{f,L_2} + Q_{\alpha/2}, \widehat{Y}_{f,L_2} + Q_{1-\alpha/2}   \right];
\end{equation}
$Q_{\alpha/2}$ and $Q_{1-\alpha/2}$ are $\alpha/2$ and $1-\alpha/2$ lower quantiles of $F_{R_f^*}$. 

We call the PI obtained through the above procedure Pertinent PI (PPI). A concrete algorithm to determine PPI will be given in \cref{Subsec:DLMFPPIAlgo}. Under mild conditions, we will show this strategy gives two benefits: (I) the estimation variability can be captured to get a more precise PI with finite samples; (II) less data is required to approximate the distribution of the predictive root compared to approximate the conditional distribution $F_{Y|X_f}$. In practice, these two benefits work together to make the PPI have a better empirical coverage rate for finite sample cases. Before making theoretical analyses of the PI (\ref{Eq:PPItoyexample}), we present a detailed algorithm in the next subsection.

\subsection{DLMF PPI algorithm}\label{Subsec:DLMFPPIAlgo}
We present a detailed \cref{Algo:DLMFPPI} to find DLMF PPI according to the specific procedure mentioned at the beginning of this section.
\begin{algorithm}
\setstretch{1.25}
\caption{DLMF pertinent prediction interval conditional on $X = x_f$}\label{Algo:DLMFPPI}
\begin{algorithmic}[1]
\Require (a) the training samples $\{X_i,Y_i\}_{i=1}^n$; (b) the new observed $x_f$; (c) $F_Z$, the distribution of the reference variable $Z$; (d) the learning rate $lr$; (e) the number of epochs; (f) the clipping parameter $m$; (g) the number of evaluations $S$ for the Monte Carlo estimation; (h) the number of bootstrap predictive roots $B$; (i) the specific optimizer, e.g., SGD, Adam and RMSProp.
\Statex
\State Simulate $\{Z_i\}_{i=1}^n$ from $F_Z$.
\State Get $\widehat{H}(\cdot, \cdot)$ by applying \cref{Algo:DLMFpoint}.
\State Simulate $\{Z_j\}_{j=1}^S$ from $F_Z$; $\{\widehat{Y}_j\}_{j=1}^{S}$ $\gets$ $\{\widehat{H}(x_f, Z_j)\}_{j=1}^S$.
\State Compute $\widehat{Y}_{f,L_2} = \frac{1}{S}\sum_{j=1}^S \widehat{Y}_j$. (The predictive root $R_f = Y_f - \widehat{Y}_{f,L_2}$; $Y_f\sim F_{Y|x_f}$)
\For{$i\in (1,2,3,\ldots, B)$ } 
\State Simulate $\{Z^*_j\}_{j=1}^n$ from $F_Z$; $\{Y^*_j\}_{j=1}^{n}$ $\gets$ $\{\widehat{H}(X_j, Z^*_j)\}_{j=1}^n$. 
\State With $\{X_i,Y^*_i, Z_i\}_{i=1}^n$, re-train the DNN with the same procedure as Step 2 to get $\widehat{H}^*(\cdot, \cdot)$.
\State Simulate another set of $\{Z_j\}_{j=1}^S$ from $F_Z$; $\{\widehat{Y}^*_j\}_{j=1}^{S}$ $\gets$ $\{\widehat{H}^*(x_f, Z_j)\}_{j=1}^S$.
\State Compute $\widehat{Y}^*_{f,L_2} = \frac{1}{S}\sum_{j=1}^S \widehat{Y}^*_j$.
\State Simulate a single $Z_f$ from $F_Z$; $Y^*_f$ $=$ $\widehat{H}(x_f, Z_f)$.
\State Compute and record the $i$-th bootstrap version predictive root $R^*_{f} = Y^*_f - \widehat{Y}^*_{f,L_2}$. 
\EndFor
\State Denote the lower $\alpha/2$ and $1-\alpha/2$ quantile of $\{R^*_{f,i}\}_{i=1}^B$ by $\hat{q}_{\alpha/2}$ and $\hat{q}_{1 - \alpha/2}$.
\State \textbf{Return:} the pertinent prediction interval conditional on $X = x_f$ with asymptotic $1-\alpha$ confidence level
\begin{equation}\label{Eq:PPI}
[\widehat{Y}_{f,L_2} +  \hat{q}_{\alpha/2},  \widehat{Y}_{f,L_2} + \hat{q}_{1 - \alpha/2} ].
\end{equation}
\Statex
\Statex \textbf{Comments:} the bootstrap version predictive roots can be easily computed parallelly to relieve the computational burden. Based on our observation from simulation studies, the final QPI, PPI or PI based on DGs will overcover the test points if the number of epochs is too large. Thus, we apply an early-stopping procedure to avoid overfitting so the generalization is not spoiled, see more details in \cref{Sec:Simulation}. 
\end{algorithmic}
\end{algorithm}
This algorithm gives PPI conditional on any future $x_f$. Instead of taking the optimal $L_2$ point prediction as the center of PPI, we can consider another center of PPI, e.g., the $L_1$ point prediction. In addition, we take a fixed design in step 6 of \cref{Algo:DLMFPPI}, i.e., we treat $\{X_j\}_{j=1}^{n}$ as fixed observations when we generate pseudo values of $\{Y_j^*\}_{j=1}^n := \{\widehat{H}(X_j, Z^*_j)\}_{j=1}^n$; $\{Z^*_j\}_{j=1}^n$ from $F_Z$. Then we re-train the DNN to approximate the inverse transformation function with the dataset $\{X_i,Y^*_i, Z_i\}_{i=1}^n$. On the other hand, we may consider the regression problem in a random-design context. This means we treat the $X$ to be random i.e., $\{Y_j^*\}_{j=1}^n = \{\widehat{H}(X^*_j, Z^*_j)\}_{j=1}^n$; $X^*_j$ is bootstrapped from the training samples $\{X_i\}_{i=1}^n$. Then, $\widehat{H}^*_{\theta}(\cdot, \cdot)$ is trained with $\{X_i^*,Y^*_i, Z_i\}_{i=1}^n$. In this paper, we focus on the fixed design setting and explore the performance of PPIs centered at $L_2$ optimal point prediction.

%{\color{red} check the simulation with random design PPI procedure}

\FloatBarrier
\subsection{Theoretical foundation of DLMF PPI}\label{Subsec:DLMFPPITheory}
In this section, we demonstrate that the PI comes from \cref{Algo:DLMFPPI} meets the conditions C1-C4. Thus, the asymptotically pertinent property of PI (\ref{Eq:PPI}) perseveres. This is the first benefit we can acquire from the DLMF PPI, i.e., the estimation variability is captured. For the benefit (II), we need a mild assumption on $\widetilde{F}_{\widehat{H}(x_f,Z)}$ which is the distribution of $\widehat{H}(x_f,Z)$ and $F_{Y|x_f}$:
\begin{itemize}
    \item A5 For large enough $n$, $\widetilde{F}_{\widehat{H}(x_f,Z)}$ and $F_{Y|x_f}$ belongs to the following class of CDFs:
$$
\mathcal{F}=\left\{F: \exists \sigma_0, u_0 \in \mathbb{R}_{+}, \forall \sigma<\sigma_0, u_0<u<M, \bar{F}(u)<\overline{F \star \phi_\sigma}(u) ; \bar{F}(-u)>\overline{F \star \phi_\sigma}(-u)\right\}.
$$
\end{itemize}
Here, $\bar{F}(u)=1-F(u)$ is the tail distribution function and $\star$ is the convolution operator; $\phi_\sigma$ is the density function of the normal distribution $N\left(0, \sigma^2\right)$. To be
more intuitive, A5 implies that the conditional distribution of $Y_f + Z$ possesses heavier tails than that of $Y_f$; $Y_f \sim \widetilde{F}_{\widehat{H}(x_f,Z)}~\text{or}~F_{Y|x_f}$ and $Z \sim N(0,\sigma^2)$. In fact, there is a large class of distributions that satisfies A5. We present the lemma below to reveal an equivalent condition with A5:

\begin{Lemma}[Lemma 5.1 of \cite{wang2021model}] Let $f(u)$ be a density function defined on $\mathcal{Y}$. We assume: (i) $f(u)$ is third-order differentiable  and $\exists~u_0>0$ such that $\left|f^{(3)}(u)\right|$ is bounded for $|u|>u_0$. (ii) $f(u)$ is convex for $|u|>u_0$, i.e., $f^{\prime \prime}(u)>0$. Then the CDF of $f$ belongs to the function class $\mathcal{F}$.
\end{Lemma}\label{Lemma:tailDistribution}

All in all, we provide the below theorem to state the advantage of PPI from \cref{Algo:DLMFPPI}:
\begin{Theorem}
   Under A1 and A3-A4, for an appropriate sequence of sets $\Omega_{n}$, such that $\mathbb{P}((\{X_i,$ $Y_i,$ $Z_i\}_{i=1}^n)$$\notin\Omega_{n}) = o(1)$, predictive roots $R_f$ and $R^*_f$ in \cref{Algo:DLMFPPI} satisfy C1-C4 under $S\to \infty$ in an appropriate rate revealed in proof for each $n$, when $n\to\infty$. Then benefit (I) is achieved. Furthermore, with the additional assumption A5, 
    \begin{equation}\label{Eq:lessdatareason}
        \sup _x\left|\widehat{F}_{\widehat{H}(x_f,Z)} \star \phi_\sigma(x)-F_{Y \mid x_f} \star \phi_\sigma(x)\right| \leq \sup _x\left|\widehat{F}_{\widehat{H}(x_f,Z)}(x)-F_{Y \mid x_f}(x)\right|~\text{with probability}~1;
    \end{equation}
    Recall that $\widehat{F}_{\widehat{H}(x_f,Z)}$ is the empirical distribution of $\{\widehat{H}(x_f,Z_i)\}_{i=1}^S$. 
\end{Theorem}
The direct consequence of \cref{Eq:lessdatareason} is that the estimation of $F_{Y \mid x_f} \star \phi_\sigma(x)$ is more accurate compared to the direct approximation of $F_{Y \mid x_f}$ with a common large enough $n$ if A5 holds. A further implication is that PPI derived with predictive roots is more accurate than the simple quantile PI which is based on the conditional distribution solely. From another perspective, to attain the same estimation precision for $F_{Y \mid x_f} \star \phi_\sigma(x)$ and $F_{Y \mid x_f}$, fewer data is demanded for the former one. The PPI and naive quantile PI will be further evaluated with simulations in \cref{Sec:Simulation}.

%%%%%%%%%%%%%%%%%%
%%%%%%%%%%%%%%%%%%
%%%%%%%%%%%%%%%%%%  Section 5 starts here
%%%%%%%%%%%%%%%%%%
%%%%%%%%%%%%%%%%%%

\section{Simulation}\label{Sec:Simulation}
We deploy simulated data to investigate the performance of DLMF and two DG-based prediction methods. We also take the Conditional Kernel Density Estimation (CKDE) method as the benchmark; the bandwidth is selected with leave-one-out maximum likelihood cross-validation; see \cite{rothfuss2019conditional} for a comprehensive introduction to the CKDE method. To make a fair comparison, we take three models from the simulation studies of \cite{zhou2023deep} to simulate data:
\begin{itemize}
    \item Model-1: $Y_i=X_{i,1}^2+\exp \left(X_{i,2}+X_{i,3} / 3\right)+\sin \left(X_{i,4}+X_{i,5}\right)+\varepsilon_i$;
    
    \item Model-2: $Y_i=X_{i,1}^2+\exp \left(X_{i,2}+X_{i,3} / 3\right)+$ $X_{i,4}-X_{i,5}+\left(0.5+X_{i,2}^2 / 2+X_{i,5}^2 / 2\right) \cdot \varepsilon_i$;
    
    \item Model-3: $Y_i=\mathbb{I}_{\{U<0.5\}} N\left(-X_{i,1}, 0.25^2\right)+\mathbb{I}_{\{U>0.5\}} N\left(X_{i,1}, 0.25^2\right), \text { where } U \sim \operatorname{Uniform}(0,1)$.
\end{itemize}
We intend to choose these three data-generating processes since they are representations of three classes of models: (1) regression model with homoscedastic error; (2) regression model with varying variance of error depending on $X$; (3) a mixture of two distributions. We take $n = 2000$ for simulation studies. To generate data from these three models, we take $X$ and $\varepsilon_i$ from $N(0,I_5)$ and $N(0,1)$ truncated to $[-5,5]$, respectively, to satisfy assumption A3.

\subsection{Conditional point prediction}\label{Subsec:CPP}
To measure different methods' performance on the condition point prediction, we take $T$ test points $\{x_{j}\}_{j=1}^{T}$ randomly from truncated $N(0,I_5)$. Pretending the model is known, we can compute $T$ true conditional mean of $Y$, i.e., $\{\mathbb{E}(Y_{j}|x_j)\}_{j=1}^{T}$. Denote the set as $\{Y_{j,L_2}\}_{j=1}^{T}$.

To predict the mean of $Y$ conditional on $X$ by $\widehat{H}$, we rely on steps 7-9 of \cref{Algo:DLMFpoint}, i.e., $\widehat{Y}_{j} = \frac{1}{S}\sum_{s=1}^{S} \widehat{H} (x_{j},Z_{s})$ for $j = 1,\ldots,T$; $Z_{s}\sim N(0, I_p)$. We denote $\mathbb{E}(\widehat{H} (x_{j},Z))$ by $\widetilde{Y}_{j}$, which is the \textit{exact} optimal point prediction based on $\widehat{H}$. $\widehat{Y}_{j}$ converges to $\widetilde{Y}_{j}$ in probability by law of large number if $S \to \infty$. We hope $\widetilde{Y}_{j}$ is available, but it may be intractable in practice. Thus, we use $\widehat{Y}_{j}$ with a large $S$ to be optimal point prediction conditional on the $j$-th test point. Similarly, we apply the same number of $S$ to find the optimal $L_2$ point prediction by deep generators $\widehat{G}_{\text{KL}}$ and $\widehat{G}_{\text{WA}}$. For the benchmark method based on CKDE, we apply the numerical integration $\int_{\mathcal{Y}} y \hat{f}_{y|x_j} dy$ with 1000 subdivisions to approximate $E(Y|x_j)$; $\hat{f}_{y|x_j}$ is the kernel conditional density estimator of $Y$ conditional on $x_j$. For each test point, we define the prediction error as:
\begin{equation}
    L_j := (Y_{j,L_2} - \widehat{Y}_{j})^2;
\end{equation}
$\widehat{Y}_{j}$ represent various optimal point predictions. Implicitly, $\widehat{Y}_{j}$ also depends on the training set $\{(X_i,Y_i,Z_i)\}_{i=1}^n$. In other words, $\widehat{Y}_{j}$ is conditional on $x_j$ and $\{(X_i,Y_i,Z_i)\}_{i=1}^n$. Thus, $L_j$ is not the exact correct measure to evaluate the point prediction conditional on $X = x_j$. To remove the conditioning of the training data set, we shall evaluate the prediction error with $R$ different training sets:
\begin{equation}
    \overline{L}_j := \frac{1}{R}\sum_{r=1}^{R}(Y_{j,L_2} - \widehat{Y}_{r,j})^2;
\end{equation}
$\widehat{Y}_{r,j}$ is the optimal point prediction based on the $r$-th training dataset $\{(X_{r,i},Y_{r,i},Z_{r,i})\}_{i=1}^n$. The lemma below explains the meaning of $\overline{L}_j$:
\begin{Lemma}\label{Lemma:upperboundofbarLj}
For each $n$, if we let $S\to\infty$ and then $R\to\infty$, $\overline{L}_j$ is an upper bound of $\left(  Y_{j,L_2} -  \widetilde{Y}_{j}  \right)^2$ in probability, i.e., $\overline{L}_j > \left(  Y_{j,L_2} -  \widetilde{Y}_{j}  \right)^2$ with probability tends to 1 conditional on each test points $x_j$. 
\end{Lemma}

To measure the overall performance of different prediction methods, we further take the average performance on all $T$ test points as the final metric:
\begin{equation}
    \widetilde{L} = \frac{1}{T}\sum_{j=1}^{T}\overline{L}_j.
\end{equation}

 To make a fair comparison between our methods and current deep generator methods, we apply the same hyperparameter setting to train $\widehat{H}$ and $\widehat{G}$. We take $n = 2000$; $T = 2000$; $S = 10000$; $R = 200$; $p = 1,3,5,10$, $m = 20$; $\text{Learning rate:}~ 0.001$; $\text{Number of epochs:}~10000$. We intend to take a small learning rate and a large number of epochs to make the slow updating to DNN. For the optimizer of the training process, \cite{arjovsky2017wasserstein} proposed using optimizer RMSProp with Wasserstein distance is more appropriate since they observed that the adversarial training process becomes unstable with an Adam-type optimizer. However, \cite{pang2020bag} argued that SGD-based optimizers are better. As far as we know, there is no clear criterion to determine which optimizer we should use for adversarial training under different cases. Thus, we consider three common optimizers, SGD, Adam and RMSProp, to train $\widehat{H}$ and $\widehat{G}$ and then evaluate the conditional prediction performance. 

For the structure of DNN, we also train the deep generator with the same structure as the inverse transformation estimator to control the effects of DNN structure. Thus, we separate the simulation studies into two groups: (a) structures of inverse transformation estimator and deep generators are [35,35] and [50], respectively; (b) structures of inverse transformation estimator and deep generators are all [35,35]. Adopted from the previous work, the discriminator has 2 hidden layers with widths [50, 25] for both groups. We should mention that training sets are different for the two groups; we intend to do this since all methods can be tested against more different training sets. 

\begin{Remark}\label{Remark:seed}
    To make our simulation studies reproducible, we apply the command torch.manual\_seed (1000) to generate 2000 test points. The training process is accomplished with \textit{PyTorch} package.
\end{Remark}

The conditional point prediction results for various methods are tabularized in \cref{Table:pointpreModel1,Table:pointpreModel2,Table:pointpreModel3}; here column with name $H_{\text{New}}$, $G_{\text{KL}}$ and $G_{\text{WA}}$ represent our proposed method and two methods based on DGs with different probability metrics, respectively. Group (a) and Group (b) represent two DNN structure groups. We can summarize several findings:
\begin{itemize}
    \item For our newly proposed method, its performance can be sustained against various algorithms, $p$ and DNN structure. On the other hand, the DG-based methods are pretty unstable no matter with KL-divergence or Wasserstein-1 distance. The instability appears on three aspects: (1) The performance of DGs varies a lot with applying different optimizer algorithms e.g., $G_{\text{KL}}$ and $G_{\text{WA}}$ work well for Model-1 with RMSProp algorithm, but the corresponding performance is much worse when Adam or SGD algorithm is applied; (2) The performance of DGs especially for method $G_{\text{WA}}$ is not consistent when the DNN structure of the deep generator changes between Group (a) and (b); (3) The performance of DGs is more sensitive to $p$ which is the dimension of the reference random variable $Z$, e.g., the performance of $G_{\text{WA}}$ on Model-3 with Group (b) structure changes a lot as $p$ changes. 

    \item Our method can achieve the lowest MSE for all three Model data. For DG methods, they can beat the benchmark method (CKDE) under the appropriate optimization algorithm and dimension $p$, but its performance is not stable and heavily relies on a good setting.

    \item The training process of our method is more stable since it is basically an optimization process w.r.t. one special MSE loss. On the other hand, the adversarial training procedure is unstable sometimes. More specifically, the $\widetilde{L}$ for methods $G_{\text{WA}}$ and $G_{\text{KL}}$ are huge for some cases. After investigation, it turns out the reason for this is that the deep generator performs pretty badly for one or two replications out of 200 replication datasets so that the final metric $\widetilde{L}$ is spoiled. 
\end{itemize}

\begin{table}[htbp]
    \centering
    \caption{Conditional point predictions of Model-1 under group (a) and (b).}
    \begin{tabular}{lccccccc}
    \toprule
     &  \multicolumn{3}{c}{Group (a)} & & \multicolumn{3}{c}{Group (b)} \\
     & $H_{\text{New}}$ & $G_{\text{KL}}$ & $G_{\text{WA}}$ &   & $H_{\text{New}}$ & $G_{\text{KL}}$ & $G_{\text{WA}}$\\
     \hline
     \multicolumn{2}{l}{SGD} \\
      $p = 1$  & 0.294& 4.052 & 7.399 & & 0.295& 3.793 & 4.709 \\
      $p = 3$   & 0.329 & 4.041 & 7.088 & &0.330 & 3.733 & 5.090 \\
      $p = 5$   & 0.361 & 4.074 & 8.237 & &0.358 & 3.753 & 5.405\\
      $p = 10$   & 0.392 & 4.119  & 7.039 &  & 0.388 & 3.765  & 8.335 \\ [5pt]
     \multicolumn{2}{l}{Adam} \\
     
      $p = 1$  & 0.494 & 0.545 & 3.780 & &0.497 & 1.478  &  4.429 \\
      $p = 3$   & 0.318 & 0.352 & 3.585 & &0.319  & 1.652 &  2.772 \\
      $p = 5$   &  0.259 & 0.264  & 3.129 & &0.259  &  1.175  & 20.243\\
      $p = 10$   & 0.223 & 0.282  &  2.428  & &0.235  & 1.259  & 1.883\\ [5pt]
     \multicolumn{2}{l}{RMSProp} \\
           $p = 1$  & 0.248 & 0.672  & 0.684 &  &0.249 & 0.707  & 0.900 \\
      $p = 3$   & 0.194 & 0.371 & 0.415   & &0.198 & 0.542 & 0.545\\
      $p = 5$   & 0.184 & 2.262 & 0.307  & &0.181 & 0.344 & 0.402\\
      $p = 10$   & 0.198  & 0.195  & 0.207   &  &0.196 & 0.257  &  0.279\\ 
    \bottomrule
    \end{tabular}
    
        \vspace{2pt}
    \raggedright
    {\footnotesize Note: The error metric $\widetilde{L}$ of using conditional kernel density estimation (i.e., the benchmark method CKDE) is around 0.767. }
\label{Table:pointpreModel1}
\end{table}

\begin{table}[htbp]
    \centering
    
    \caption{Conditional point predictions of Model-2 under group (a) and (b).}
    \begin{tabular}{lccccccc}
    \toprule
     &  \multicolumn{3}{c}{Group (a)} & & \multicolumn{3}{c}{Group (b)} \\
     & $H_{\text{New}}$ & $G_{\text{KL}}$ & $G_{\text{WA}}$ &   & $H_{\text{New}}$ & $G_{\text{KL}}$ & $G_{\text{WA}}$\\
     \hline
     \multicolumn{2}{l}{SGD} \\
      $p = 1$  & 0.309 & 3.931& 10.390 & & 0.292 & 3.827 & 82.971\\
      $p = 3$   & 0.298 & 4.009 & 11.096 & & 0.285  &  3.762 &  56644 \\
      $p = 5$   & 0.296 & 4.036 & 40.388 & & 0.281 & 3.801  & 12843\\
      $p = 10$   & 0.294 & 4.116  & 182.343 &  & 0.280  &  3.812 & 11378 \\ [5pt]
     \multicolumn{2}{l}{Adam} \\
     
      $p = 1$  & 1.608 & 1.838 & 3558 & & 1.572 & 1.836 & 14322 \\
      $p = 3$   & 0.832 & 1.105 & 8.480 & & 0.843  &  1.549 & 43.483 \\
      $p = 5$   &  0.604 & 0.820 & 43.853 & & 0.591  & 1.166    & 43.837 \\
      $p = 10$   & 0.412 & 0.495  &  5.523  & & 0.422  & 0.817  & 14.496 \\ [5pt]
     \multicolumn{2}{l}{RMSProp} \\
           $p = 1$  &0.960 & 1.767  & 1.910 &  & 0.973 & 1.620 & 2.326 \\
      $p = 3$   & 0.601 & 1.049 & 1.248   & & 0.597 & 0.964 & 1.263 \\
      $p = 5$   & 0.484 & 0.779 & 0.908 & & 0.479 & 0.727 & 0.903 \\
      $p = 10$   &0.365 & 0.463  & 0.598   &  &  0.352 & 0.494  &  0.508\\ 
    \bottomrule
    \end{tabular}
        
    \vspace{2pt}
    \raggedright
    {\footnotesize Note: The error metric $\widetilde{L}$ of using conditional kernel density estimation (i.e., the benchmark method CKDE) is around 1.210.}
\label{Table:pointpreModel2}
\end{table}

\begin{table}[htbp]
    \centering
   
    \caption{Conditional point predictions of Model-3 under group (a) and (b).}
    \begin{tabular}{lccccccc}
    \toprule
     &  \multicolumn{3}{c}{Group (a)} & & \multicolumn{3}{c}{Group (b)} \\
     & $H_{\text{New}}$ & $G_{\text{KL}}$ & $G_{\text{WA}}$ &   & $H_{\text{New}}$ & $G_{\text{KL}}$ & $G_{\text{WA}}$\\
     \hline
     \multicolumn{2}{l}{SGD} \\
      $p = 1$  & 0.004 & 0.092& 0.999 & & 0.004 & 0.125 & 1.043\\
      $p = 3$   & 0.003 & 0.034 & 1.186 & & 0.003  &  0.079 &  1.142 \\
      $p = 5$   & 0.002 & 0.041 & 1.244 & & 0.002 & 0.076  & 42.143\\
      $p = 10$   & 0.002 & 0.071  & 15.491 &  & 0.002  &  0.097 & 199.893 \\ [5pt]
     \multicolumn{2}{l}{Adam} \\
     
      $p = 1$  & 0.066 & 0.220 & 2.139e+07 & & 0.064 & 0.163 & 7.780e+07 \\
      $p = 3$   & 0.035 & 0.278 & 1.429e+04 & & 0.032 &  0.184 & 2.094e+04 \\
      $p = 5$   &  0.025 & 0.268 & 1.618e+07 & & 0.023 & 0.1922   & 1.741e+08 \\
      $p = 10$   & 0.019 & 0.257  &  1.004e+05  & & 0.016  & 0.195  & 1.904e+04 \\ [5pt]
     \multicolumn{2}{l}{RMSProp} \\
           $p = 1$  &0.042 & 0.104  & 547.273 &  & 0.043 & 0.068 & 0.500 \\
      $p = 3$   & 0.041 & 0.104 & 488.320 & & 0.039 & 0.085 & 0.544 \\
      $p = 5$   & 0.039 &  0.156 & 4142.580 & & 0.035 & 0.090 & 87.538 \\
      $p = 10$   &0.035 & 0.181  & 0.796   &  &  0.032 & 0.126  &  1.295\\ 
    \bottomrule
    \end{tabular}
        
    \vspace{2pt}
    \raggedright
    {\footnotesize Note: The error metric $\widetilde{L}$ of using conditional kernel density estimation (i.e., the benchmark method CKDE) is pretty large and even the NaN values would be generated.}
    \label{Table:pointpreModel3}
\end{table}

\FloatBarrier
 
\subsection{Conditional prediction interval}\label{Subsec:conditionalPI}
To measure the performance of different methods' PIs, we still take $T$ test points $\{x_{j}\}_{j=1}^{T}$ randomly from truncated $N(0,I_5)$ according to the same random seed in \cref{Remark:seed}. As commented in \cref{Algo:DLMFPPI}, we observed that the PI from DG and our methods overcover these test points. Therefore, we use a validation dataset with size $V$ to do early-stopping. After each $O$ number of epochs, we evaluate the performance of different PIs on the validation dataset. If the empirical coverage rate is larger than the nominal confidence level $1-\alpha$, the training will be terminated; otherwise, the training procedure will go through all epochs. 

Similar to the analysis of conditional point prediction in \cref{Subsec:CPP}, the direct conditional PIs returned by our and DG methods are also conditional on training dataset $\{(X_i,Y_i)\}_{i=1}^n$ and the test point $X_f = x_f$. To simplify notation, we denote all PIs considered in this section by $\widehat{\mathcal{I}}$, we first consider the coverage probability given the training dataset and a specific test point:
\begin{equation}\label{Eq:PI3}
    P(Y_f \in \widehat{\mathcal{I}} | x_f, \{(X_i,Y_i)\}_{i=1}^n).
\end{equation}
Since the simulation model is known to us, the conditional probability (\ref{Eq:PI3}) can be approximated with any level of accuracy by Monte Carlo simulation. In short, we can simulate $S^{\prime}$ number of pseudo values $\{Y^*_{j}\}_{j=1}^{S^{\prime}}$ conditional on $X_f = x_f$ through simulation $S^{\prime}$ number of errors. Then, we can use the empirical coverage rate of $\widehat{\mathcal{I}}$ on $\{Y^*_{j}\}_{j=1}^{S^{\prime}}$ to estimate probability (\ref{Eq:PI3}). Observing that
\begin{equation}\label{Eq:PI2}
    P(Y_f \in \widehat{\mathcal{I}} | x_f)
\end{equation}
is equal to $\mathbb{E}_{\{(X_i,Y_i)\}_{i=1}^n}(P(Y_f \in \widehat{\mathcal{I}} | x_f, \{(X_i,Y_i)\}_{i=1}^n))$ by tower property. Thus, to remove the conditioning on training datasets in practice, we can approximate probability (\ref{Eq:PI2}) by taking average of $P(Y_f \in \widehat{\mathcal{I}} | x_f, \{(X_i,Y_i)\}_{i=1}^n)$ on $R$ different training sets; each $P(Y_f \in \widehat{\mathcal{I}} | x_f, \{(X^{(r)}_i,Y^{(r)}_i)\}_{i=1}^n)$ can be approximated by simulation method for $r = 1,\ldots, R$. If we further taking average of $P(Y_f \in \widehat{\mathcal{I}} | x_f)$ w.r.t. all test point, i.e., $\mathbb{E}_{X}(P(Y_f \in \widehat{\mathcal{I}} | X))$, we can get the unconditional probability: 
\begin{equation}\label{Eq:PI1}
    P(Y_f \in \widehat{\mathcal{I}}).
\end{equation}
In our simulation studies, we present the empirical coverage rate for different PIs after taking the average w.r.t. $T$ test points and $R$ training datasets, i.e., $\frac{1}{T}\sum_{j=1}^T P(Y_f \in \widehat{\mathcal{I}} | x_{j})~\text{for~} j = 1,\ldots, T$. Each $P(Y_f \in \widehat{\mathcal{I}} | x_{j})$ can be approximated by the above simulation method. This average empirical coverage rate can be thought of as an estimation of probability (\ref{Eq:PI1}). To avoid confusion, we denote the estimated probability (\ref{Eq:PI1}), (\ref{Eq:PI2}) and (\ref{Eq:PI3}) by $CV_{1}$, $CV_{2}$ and $CV_{3}$, respectively. In other words, $CV_{1}$, $CV_{2}$ and $CV_{3}$ represent the coverage rate of different PIs under different conditioning levels. 

\begin{Remarknn}[Different levels of conditioning]
    There are three probabilities related to the coverage of PI. It is easy to see that the conditional probability (\ref{Eq:PI3}) is the strongest one, which means the $1-\alpha$ coverage under probability (\ref{Eq:PI3}) implies the same coverage level under probabilities (\ref{Eq:PI2}) and (\ref{Eq:PI1}), see the work of \cite{wang2021model} for more discussions.
\end{Remarknn}

According to the simulation results from \cref{Subsec:CPP}, the optimization algorithm RMSProp works best for our and DG methods generally speaking. Thus, we take the RMSProp algorithm throughout simulation studies for conditional PI. For hyperparameter setting, we take $T = 2000$; $S = 10000$; $R = 200$; $m = 20$; $V = 200$; $O = 500$; $B = 200$; $\alpha = 0.05$ (i.e., nominal confidence level is 95$\%$); $\text{Learning rate:}~ 0.001$; $\text{Number of epochs:}~5000$. Since the construction of PI is a harder task than $L_2$ point prediction, we take $p = 5, 10, 15, 20, 25$. In addition, we take three different sample sizes $n = 200, 500, 2000$ to create short and large sample situations. For the DNN structure of our and DG methods, we set $\widehat{H}$, $\widehat{G}_{\text{KL}}$ and $\widehat{G}_{\text{WA}}$ to be a neural network with one hidden layer and 50 neurons. The discriminator has 2 hidden layers with widths [50, 25]. 

We consider 4 types of PI: (1) QPI based on the DLMF prediction algorithm; (2) PPI based on the DLMF prediction algorithm; (3) PI-KL: QPI based on $\widehat{G}_{\text{KL}}$; (4) PI-WA: QPI based on $\widehat{G}_{\text{WA}}$. The $CV_{1}$ for different PIs are presented in \cref{Table:fixpvaryingnM1,Table:fixpvaryingnM2,Table:fixpvaryingnM3}. Besides $CV_{1}$, we also consider the sample standard deviation $\hat{\sigma}_{\text{PI}}$ of 2000 $CV_{2}$; $\hat{\sigma}_{\text{PI}}$ can measure the spread of the conditional probability (\ref{Eq:PI2}). A superior method shall be able to give a PI with $CV_{1}$ close to a nominal confidence level and $\hat{\sigma}_{\text{PI}}$ is as small as possible.  We also present the average length (AL) of different PIs. The AL of one specific type of PI is also computed as a double average w.r.t. 200 training dataset and 2000 test points. Similarly, the sample standard deviation $\hat{\sigma}_{\text{Len}}$ of 2000 PIs' length after taking average w.r.t. 200 training samples is also provided along with AL. If two PIs have a close $CV_{1}$, the one with a small AL and $\hat{\sigma}_{\text{Len}}$ will be better in practice.

From simulations, all PIs work well when the sample size is large, i.e., $n = 2000$. However, for small sample situations, i.e., $n = 200$, the performance is different. Take Model-2 with 200 samples as an example, by tuning the dimension of the reference variable $p$, QPI and PPI achieve the best performance regarding $CV_{1}$ with $p = 10$; PI-KL and PI-WA achieve the best performance with $p = 20$. However, the $CV_{1}$ of PI-KL and PI-WA is still below the nominal level. On the other hand, the PPI and QPI are better and can achieve $CV_{1}$ closer to the nominal level. Generally speaking, PPI is the best one. Its $CV_{1}$ is closest to $95\%$ and $\hat{\sigma}_{\text{PI}}$ is smallest meanwhile for Model-1 and Model-2 data when $p = 5$ and $p = 10$ respectively. For Model-3 data, PPI with $p=5$ slightly overcovers the test points, but its $\hat{\sigma}_{\text{PI}}$ is still the smallest.

\begin{figure}[htbp]
    \centering
    \includegraphics[scale = 0.3]{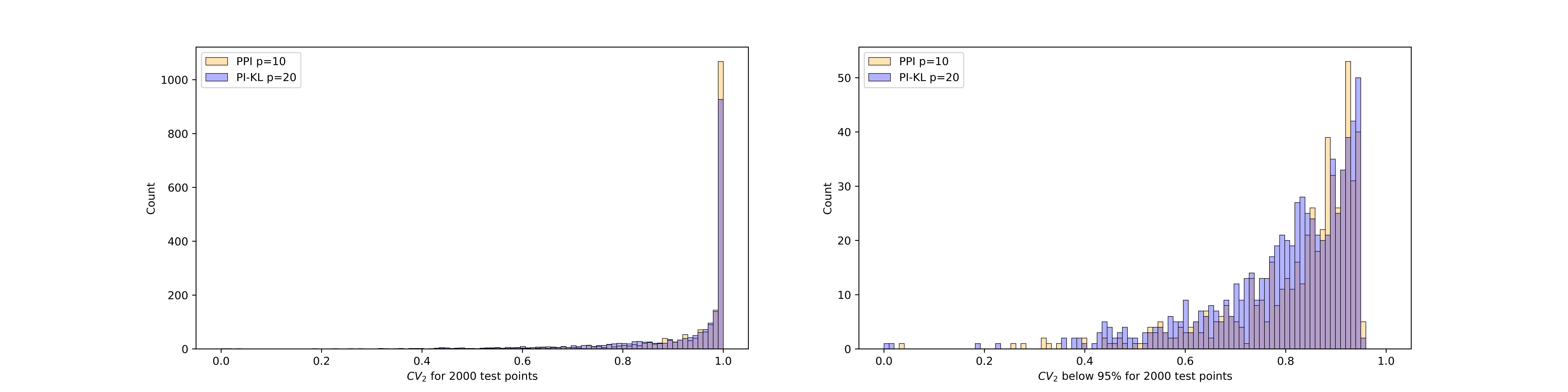}
    \caption{Histograms of undercoverage $CV_{2}$ ($CV_{2}$ less than nominal level 95$\%$ ) and all $CV_{2}$ of PPI and PI-KL.}
    \label{Fig:PPIvsPIKL}
\end{figure}

\begin{figure}[htbp]
    \centering
    \includegraphics[scale = 0.3]{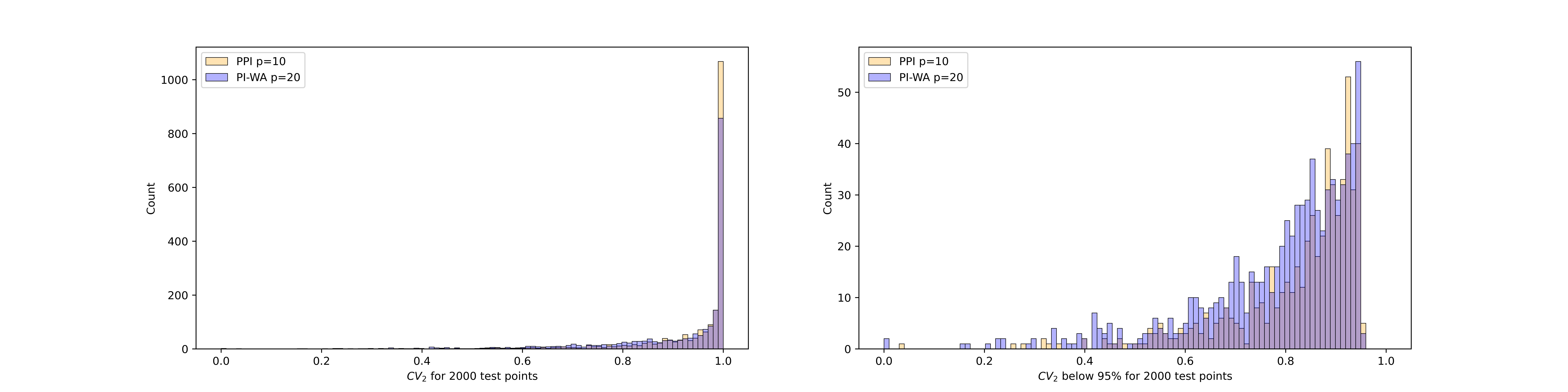}
    \caption{Histograms of undercoverage $CV_{2}$ ($CV_{2}$ less than nominal level 95$\%$ ) and all $CV_{2}$ of PPI and PI-WA.}
    \label{Fig:PPIvsPIWA}
\end{figure}

\begin{figure}[htbp]
    \centering
    \includegraphics[scale = 0.3]{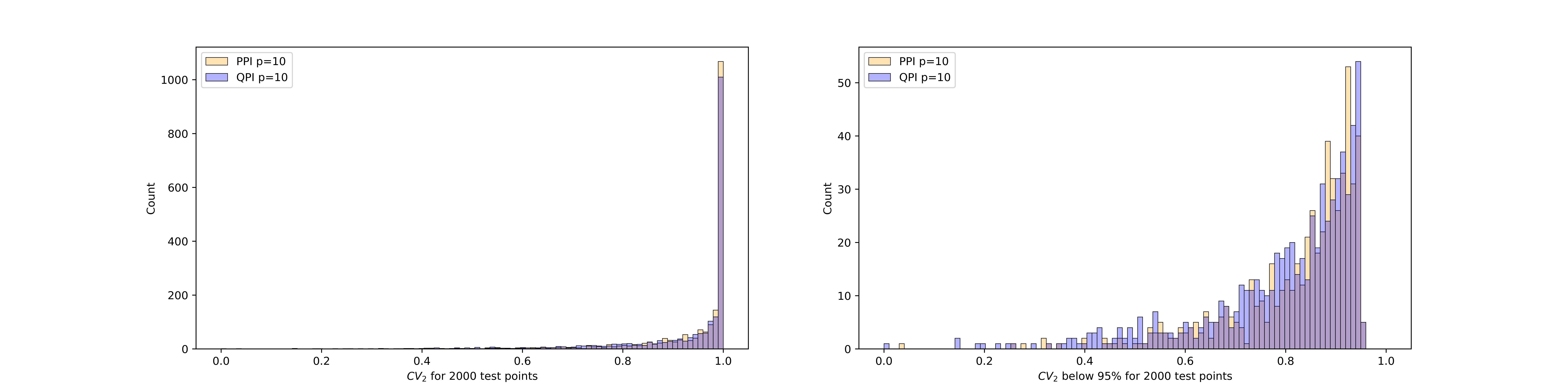}
    \caption{Histograms of undercoverage $CV_{2}$ ($CV_{2}$ less than nominal level 95$\%$ ) and all $CV_{2}$ of PPI and QPI.}
    \label{Fig:PPIvsQPI}
\end{figure}

To get a close look at the performance of different PIs according to $CV_{2}$ when $n = 200$, we take Model-2 case as an example. We consider histograms of undercoverage $CV_{2}$ out of 2000 values ($CV_{2}$ less than nominal level 95$\%$ ) and all $CV_{2}$ between PPI and the other three PIs. These histograms are presented in \cref{Fig:PPIvsPIKL,Fig:PPIvsPIWA,Fig:PPIvsQPI}. Comparing PPI and PI-KL/PI-WA/QPI, the PPI has a higher bar near the coverage rate 1 which indicates that the 2000 $CV_{2}$ of PPI are more concentrated around a high coverage level. For the left tail part in which $CV_{2}$ is less than 0.95, bars of PI-KL/PI-WA/QPI almost cover the bars of PPI which shows PI-KL/PI-WA/QPI gives more low $CV_{2}$.

\section{Empirical studies}\label{Sec:empiricalcase}

In this section, we deploy the empirical studies to further compare our methods and DNN-based counterparts. As revealed in the \cref{Sec:Simulation}, our DLMF prediction method is more accurate and stable for optimal $L_2$ point predictions. Moreover, the specifically designed PPI covers true future values with a rate closer to the nominal level compared to PI variants returned by deep generator methods when the sample is short. Here, we apply different techniques to the wine quality dataset \cite{cortez2009modeling}. This dataset is available at UCI machine learning repository \url{https://archive.ics.uci.edu/dataset/186/wine+quality}. Two datasets (size 1599 and 4898, respectively) about red and white wine are included. For each dataset, there are eleven physicochemical quantitative variables as predictors and a quantitative variable which is a score between 0 and 10 to measure the wine quality. We hope to make a model to forecast the quality of a specific type of wine given values of 11 predictors so that such a model can be used by oenologists to improve the wine quality. 

To test the performance of estimated inverse transformation $\widehat{H}$ in DLMF method and two deep generators $\widehat{G}_{\text{KL}}$ and $\widehat{G}_{\text{WA}}$ especially for short data, we randomly take $n = 199$ and $n = 195$ data points from red and white wine datasets which correspond 1/8 and 1/25 of whole datasets, respectively. Then, we take the $80\%$ and $20\%$ of left data to be test data and validation data, respectively. To make our study reproducible, we apply the function \textit{train$\_$test$\_$split} from \textit{sklearn.model$\_$selection} to do such splitting with \textit{ random$\_$state=1}. The validation data is used to stop the training procedure early as described in \cref{Subsec:conditionalPI}. To measure the point prediction accuracy of different methods, we consider the mean square prediction error (MSPE) for different methods; to measure the performance of different PIs, we consider probability (\ref{Eq:PI3}) which is conditional on the training sample $\{(X_i,Y_i)\}_{i=1}^n$ and a new test point $x_f$. To simplify the presentation, we compute the average of the probability (\ref{Eq:PI3}) w.r.t. all test points for red and white wine test datasets. In other words, we try to evaluate the average coverage rate of the PI conditional on some specific training sample $\{(X_i,Y_i)\}_{i=1}^n$. However, we should mention that we can not apply the Monte Carlo simulation to estimate the probability (\ref{Eq:PI1}), (\ref{Eq:PI2}) and (\ref{Eq:PI3}) since the true data-generating process is unknown for real-world data. Thus, the average coverage rate of different PIs considered in this section is computed by $\sum_{j=1}^{n_{\text{test}}}{\mathbbm{1}(y_j\in \widehat{\mathcal{I}}_{j})}/n_{\text{test}}$; $\mathbbm{1}$ is the indicator function; $n_{\text{test}}$ is the size of test datasets of red and white wine; $y_j$ is the $j$-th test point; $\widehat{\mathcal{I}}_{j}$ represents different PIs. We also present the corresponding average length of the prediction interval, which can be computed by $\sum_{j=1}^{n_{\text{test}}}\text{Len}_{j}/n_{\text{test}}$; $\text{Len}_{j}$ represents the length of $\widehat{\mathcal{I}}_{j}$ for different approaches.

We apply the same setting as the simulation studies, i.e., we set $\widehat{H}$, $\widehat{G}_{\text{KL}}$ and $\widehat{G}_{\text{WA}}$ to be a neural network with one hidden layer and 50 neurons. The discriminator has 2 hidden layers with widths [50, 25]. We set the learning rate as 0.001 and the number of epochs as 5000. We consider the dimension of reference variable $p = 5,10,15,20,25$ and take the optimization algorithm RMSProp. All other hyperparameters are the same as those used in the simulation studies. The empirical results of point predictions and prediction intervals are presented in \cref{Table:empiricalpoint,Table:empiricalPI}, respectively. 

For the point prediction, the DLMF method gives optimal results with different $p$ values for two datasets. Moreover, the MSPE of DLMF varies mildly for different $p$. On the other hand, the MSPE of two deep-generator methods changes a lot when $p$ varies. For the PI, PPI with $p = 25$ is the optimal choice for two datasets according to the overall performance based on the average coverage rate and length. 
%In other words, considering different PIs that have an average coverage rate close to $95\%$, the PPI will have the shortest average length. 

\begin{table}[htbp]
    \centering
   
    \caption{MSPE of point predictions of different methods with the wine datasets.}
    \begin{tabular}{lccccccc}
    \toprule
     &  \multicolumn{3}{c}{Red wine} & & \multicolumn{3}{c}{White wine} \\
     & $H_{\text{New}}$ & $G_{\text{KL}}$ & $G_{\text{WA}}$ &   & $H_{\text{New}}$ & $G_{\text{KL}}$ & $G_{\text{WA}}$\\
     \hline
           $p = 5$  &0.552 & 0.657  & 0.756 &  & 0.865 & 1.048  &1.162  \\
      $p = 10$   & 0.541 & 0.645 & 0.544 & & \textbf{0.776} & 0.924  & 0.958 \\
      $p = 15$   & 0.516 &  0.746 & 0.596 & & 0.860 & 1.237  & 1.125 \\
      $p = 20$   & 0.531 & 0.769  & 0.525  & & 0.976  & 1.464 & 1.884  \\ 
      $p = 25$   & \textbf{0.495} & 0.625  & 0.544   &  & 0.923 & 1.127  & 1.647\\ 
    \bottomrule
    \end{tabular}
        
    \vspace{2pt}
    \raggedright
    {\footnotesize Note: $H_{\text{New}}$, $G_{\text{KL}}$ and $G_{\text{WA}}$ represent our DLMF and two deep generator methods. The optimal results are marked in bold.}
    \label{Table:empiricalpoint}
\end{table}

\begin{table}[htbp]
    \centering
    \caption{Empirical coverage rate of different PIs with wine datasets.}
    \begin{tabular}{lccccc}
    \toprule
    & \multicolumn{2}{c}{Red wine} & \multicolumn{2}{c}{White wine} \\
      &  \multicolumn{1}{c}{Coverage rate}  & \multicolumn{1}{c}{Average length} & \multicolumn{1}{c}{Coverage rate}   & \multicolumn{1}{c}{Average length}  &\\
\midrule
p = 5 &  & &  &  \\
QPI   &  0.730 &  1.477 &  0.257 &  0.712 \\
PPI   &  0.798 &  1.815 &  0.688 &  1.911 \\
PI-KL &  0.793 &  2.243 &  0.738 &  1.900 \\
PI-WA &  0.867 &  2.413 &  0.577 &  1.559 \\ 
p = 10 &  & &  &  \\
QPI   &  0.920 &  2.346 &  0.396 &  1.017 \\
PPI   &  0.930 &  2.650 &  0.707 &  1.926 \\
PI-KL &  0.971 &  3.695 &  0.021 &  1.957 \\
PI-WA &  0.942 &  3.402 &  0.958 &  4.869 \\ 
p = 15 &  & &  &  \\
QPI   &  0.969 &  3.200 &  0.659 &  1.775 \\
PPI   &  0.973 &  3.530 &  0.835 &  2.447 \\
PI-KL &  0.950 &  3.285 &  0.988 &  5.466 \\
PI-WA &  0.952 &  3.369 &  0.991 &  6.310 \\ 
p = 20 &  & &  &  \\
QPI   &  0.933 &  3.002 &  0.704 &  1.944 \\
PPI   &  0.944 &  3.274 &  0.859 &  2.597 \\
PI-KL &  0.961 &  3.174 &  0.985 &  4.745 \\
PI-WA &  0.962 &  3.303 &  0.855 &  2.266 \\
p = 25 &  & &  &  \\
QPI   &  0.937 &  2.631 &  0.824 &  2.890 \\
PPI   &  \textbf{0.942} &  \textbf{2.859} &  \textbf{0.944} &  \textbf{3.466} \\
PI-KL &  0.461 &  2.916 &  0.996 &  6.947 \\
PI-WA &  0.920 &  2.597 &  0.867 &  2.447 \\ 
    \bottomrule
    \end{tabular}
\label{Table:empiricalPI}

  \raggedright
    \vspace{2pt}
    
     {\footnotesize \textit{Note:} For this table, ``Coverage rate '' represents the average conditional empirical coverage rate of different PIs for all test points under the probability (\ref{Eq:PI3}) with the red or white dataset; Average length represents the average length of the PIs for all test points with the red or white dataset. According to the coverage rate and length, we highlight the optimal type of PI for two datasets in bold.} 
\end{table}

\FloatBarrier

\section{Conclusion}\label{Sec:Conclu}

In this paper, we propose the Deep Limit Model-free (DLMF) prediction framework which provides a solution to make point predictions and build related prediction intervals in the regression context based on a DNN without any underlying model assumptions. Being different from other DNN-based counterparts, our method is motivated by the Model-free prediction principle and the training procedure is designed with a specific loss function. The theoretical foundation of our method relies on the universal approximation ability of DNN and the noise-outsourcing lemma. In practice, the DLMF prediction method is more accurate on optimal $L_2$ point predictions compared to other deep generator methods and the classical kernel-based method. Moreover, it is also more robust against the choice of the optimization algorithm, the DNN structure, and the dimension of the reference random variable. Beyond the superiority of the DLMF method on point predictions, it can capture the estimation variability so that the corresponding prediction interval works well even for short samples.

\FloatBarrier
\section*{Acknowledgement}
This research was done using computational services provided by the OSG Consortium \cite{osg07,osg09,osg2,osg1}, which is supported by the National Science Foundation awards 2030508 and 1836650. The research of the second author was partially supported by NSF grant DMS 24-13718.

\clearpage
\begin{appendices}

\section{\textsc{Complementary tables and figures}}\label{Appendix:A}

\FloatBarrier
    \begin{table}[htbp]
    \centering
    \caption{Simulation results of conditional PIs for Model-1 with varying $n$ and $p$.}
    \begin{tabular}{lccccccc}
    \toprule
      &  \multicolumn{1}{c}{$CV_{1}$}  & \multicolumn{1}{c}{AL} & \multicolumn{1}{c}{$CV_{1}$}   & \multicolumn{1}{c}{AL}  & \multicolumn{1}{c}{$CV_{1}$}   & \multicolumn{1}{c}{AL}\\
\midrule
p = 5 &   \multicolumn{2}{c}{n = 200}  &   \multicolumn{2}{c}{n = 500} &   \multicolumn{2}{c}{n = 2000}\\
QPI & 0.916(0.123) & 4.783(0.787) & 0.952(0.078) & 4.860(0.789) & 0.738(0.097) & 2.439(0.553) \\
PPI & 0.948(0.090) & 5.381(1.085) & 0.962(0.067) & 5.242(1.043) & 0.745(0.099) & 2.540(0.640)\\
PI-KL & 0.881(0.137) & 4.186(0.837)& 0.897(0.125) & 4.386(0.885) & 0.920(0.075) & 4.096(0.784)\\
PI-WA & 0.867(0.134) & 4.066(0.795)& 0.920(0.120) & 4.900(1.439) & 0.922(0.069) & 4.181(0.825) \\[2pt]
    p = 10 &   &   & \\
QPI & 0.926(0.154) & 5.337(0.527) & 0.959(0.071) & 5.289(0.565) & 0.800(0.070) & 2.758(0.505) \\
PPI & 0.950(0.126) & 5.754(0.784) & 0.971(0.054) & 5.569(0.795) & 0.810(0.069) & 2.853(0.612)\\
PI-KL & 0.932(0.117) & 4.999(0.547) & 0.945(0.091) & 5.060(0.541) & 0.955(0.041) & 4.406(0.633)\\
PI-WA & 0.942(0.115) & 5.166(0.519) & 0.955(0.079) & 5.254(0.980) & 0.952(0.064) & 4.726(0.911)  \\[2pt]
    p = 15 &   &   & \\
QPI & 0.912(0.179) & 5.729(0.471) & 0.958(0.080) & 5.378(0.534) & 0.939(0.067) & 4.265(0.502) \\
PPI & 0.943(0.140) & 6.010(0.670) & 0.971(0.059) & 5.618(0.714) & 0.941(0.061) & 4.341(0.616)  \\
PI-KL & 0.939(0.131) & 5.829(0.612) & 0.947(0.098) & 5.193(0.625) & 0.954(0.052) & 4.453(0.455)  \\
PI-WA & 0.927(0.131) & 5.196(0.613) & 0.969(0.074) & 5.775(0.967) & 0.952(0.065) & 4.758(0.982)  \\[2pt]
    p = 20 &   &   & \\
QPI & 0.862(0.207) & 5.165(0.286) & 0.974(0.084) & 6.504(0.542)  & 0.952(0.065) & 4.646(0.329)\\
PPI & 0.885(0.186) & 5.328(0.475) & 0.986(0.051) & 6.713(0.757) & 0.955(0.061) & 4.719(0.477) \\
PI-KL & 0.892(0.163) & 5.198(0.592) & 0.958(0.089) & 5.564(0.442)  & 0.949(0.086) & 4.780(0.445) \\
PI-WA & 0.893(0.181) & 5.113(0.308) & 0.964(0.077) & 5.556(0.680) & 0.958(0.068) & 5.144(1.068) \\
    p = 25 &   &   & \\
QPI & 0.875(0.187) & 5.691(0.399) & 0.975(0.094) & 7.432(0.426) & 0.952(0.072) & 4.592(0.303) \\
PPI & 0.903(0.165) & 5.842(0.547) & 0.986(0.069) & 7.600(0.676) & 0.954(0.066) & 4.646(0.444)\\
PI-KL & 0.882(0.169) & 5.072(0.475) & 0.967(0.085) & 6.188(0.638) & 0.961(0.070) & 4.961(0.632)\\
PI-WA & 0.894(0.168) & 5.434(0.473) & 0.970(0.080) & 6.234(0.728) & 0.966(0.067) & 5.335(0.884)\\
    \bottomrule
    \end{tabular}
     \raggedright
    \vspace{2pt}
    
     {\footnotesize\textit{Note:} For this and subsequent two tables, $CV_{1}$ represents the unconditional empirical coverage rate for different PIs under the probability (\ref{Eq:PI1}); AL represents the average length of one specific PI which is also computed as a double average w.r.t. 200 training dataset and 2000 test points; the values inside the parentheses are $\hat{\sigma}_{\text{PI}}$ and $\hat{\sigma}_{\text{Len}}$, respectively.}
\label{Table:fixpvaryingnM1}
\end{table}

\begin{table}[htbp]
    \centering
    \caption{Simulation results of conditional PIs for Model-2 with varying $n$ and $p$.}
    \begin{tabular}{lccccccc}
    \toprule
      &  \multicolumn{1}{c}{$CV_{1}$}  & \multicolumn{1}{c}{AL} & \multicolumn{1}{c}{$CV_{1}$}   & \multicolumn{1}{c}{AL}  & \multicolumn{1}{c}{$CV_{1}$}   & \multicolumn{1}{c}{AL}\\
\midrule
p = 5 &   \multicolumn{2}{c}{n = 200}  &   \multicolumn{2}{c}{n = 500} &   \multicolumn{2}{c}{n = 2000}\\
QPI & 0.861(0.170) & 5.487(1.054) & 0.927(0.110) & 6.734(1.463) & 0.787(0.177) & 3.621(0.855) \\ 
PPI & 0.893(0.139) & 6.208(1.384) & 0.941(0.095) & 7.258(1.808) & 0.789(0.173) & 3.728(0.959)\\
PI-KL & 0.842(0.193) & 5.496(0.861) & 0.869(0.157) & 5.434(1.218) & 0.913(0.104) & 5.670(2.282)  \\
PI-WA & 0.852(0.181) & 5.439(0.907) & 0.882(0.150) & 5.970(2.030) & 0.899(0.105) & 5.365(1.996) \\[2pt]
    p = 10 &   &   & \\
QPI & 0.928(0.129) & 7.497(0.720) & 0.949(0.094) & 8.194(0.950) & 0.855(0.157) & 4.474(0.817)  \\
PPI & 0.944(0.105) & 8.103(1.072) & 0.961(0.076) & 8.623(1.325) & 0.855(0.154) & 4.546(0.953)\\
PI-KL & 0.900(0.133) & 6.701(0.835) & 0.925(0.119) & 6.806(0.933) & 0.928(0.099) & 5.882(1.403) \\
PI-WA & 0.898(0.146) & 6.757(0.719) & 0.933(0.116) & 7.545(1.340) & 0.934(0.100) & 6.199(1.880) \\[2pt]
    p = 15 &   &   & \\
QPI & 0.915(0.137) & 7.408(0.669) & 0.945(0.097) & 7.430(0.949) & 0.915(0.123) & 5.895(0.647) \\
PPI & 0.930(0.119) & 7.760(0.936) & 0.953(0.085) & 7.749(1.172) & 0.916(0.121) & 5.971(0.807)  \\
PI-KL & 0.909(0.136) & 7.427(0.817) & 0.949(0.095) & 8.082(1.068)  & 0.943(0.089) & 6.556(1.491) \\
PI-WA & 0.901(0.137) & 6.797(0.687) & 0.950(0.095) & 7.972(1.312) & 0.947(0.088) & 6.778(1.541)  \\[2pt]
    p = 20 &   &   & \\
QPI & 0.879(0.172) & 6.726(0.485) & 0.959(0.085) & 8.830(0.683) & 0.940(0.102) & 6.849(0.562) \\
PPI & 0.893(0.154) & 6.941(0.702) & 0.966(0.073) & 9.100(0.950) & 0.942(0.097) & 6.925(0.759)\\
PI-KL & 0.923(0.126) & 7.799(0.842) & 0.954(0.087) & 8.311(0.861) & 0.946(0.093) & 6.806(1.097) \\
PI-WA & 0.910(0.140) & 7.402(0.698) & 0.945(0.099) & 8.011(0.800) & 0.946(0.092) & 6.804(1.534)\\
   p = 25 &   &   & \\
QPI & 0.871(0.172) & 7.020(0.287) & 0.961(0.088) & 9.633(0.645) & 0.946(0.099) & 7.296(0.475)\\
PPI & 0.884(0.160) & 7.189(0.548) & 0.967(0.078) & 9.881(0.938) & 0.948(0.095) & 7.370(0.695)\\
PI-KL & 0.907(0.142) & 7.370(0.618) & 0.954(0.090) & 8.670(0.813) & 0.945(0.093) & 6.915(1.009)  \\
PI-WA & 0.897(0.151) & 7.071(0.510) & 0.960(0.081) & 8.514(0.942) & 0.944(0.097) & 7.117(1.491) \\
    \bottomrule
    \end{tabular}
\label{Table:fixpvaryingnM2}
\end{table}

\begin{table}[htbp]
    \centering
    \caption{Simulation results of conditional PIs for Model-3 with varying $n$ and $p$.}
    \begin{tabular}{lccccccc}
    \toprule
      &  \multicolumn{1}{c}{$CV_{1}$}  & \multicolumn{1}{c}{AL} & \multicolumn{1}{c}{$CV_{1}$}   & \multicolumn{1}{c}{AL}  & \multicolumn{1}{c}{$CV_{1}$}   & \multicolumn{1}{c}{AL}\\
\midrule
p = 5 &   \multicolumn{2}{c}{n = 200}  &   \multicolumn{2}{c}{n = 500} &   \multicolumn{2}{c}{n = 2000}\\
QPI &  0.967(0.106) & 3.086(0.728) & 0.954(0.121) & 3.169(0.652) & 0.651(0.283) & 1.931(0.748)\\
PPI & 0.975(0.060) & 3.159(0.836) & 0.952(0.116) & 3.193(0.707) & 0.649(0.284) & 1.940(0.774) \\
PI-KL & 0.857(0.191) & 2.697(0.653) & 0.905(0.136) & 2.535(0.883) & 0.806(0.195) & 2.021(0.989) \\
PI-WA &  0.983(0.058) & 3.173(1.134) & 0.960(0.072) & 2.652(1.167) & 0.942(0.021) & 2.529(1.324) \\[2pt]
    p = 10 &   &   & \\
QPI   & 0.964(0.126) & 4.207(0.162) & 0.957(0.117) & 3.751(0.368) & 0.902(0.205) & 3.031(0.369)\\
PPI &  0.967(0.110) & 4.281(0.374) & 0.957(0.115) & 3.778(0.485) & 0.897(0.206) & 3.032(0.432)\\
PI-KL & 0.965(0.133) & 4.083(0.182) & 0.868(0.192) & 3.077(0.602) & 0.753(0.221) & 1.930(0.272) \\
PI-WA & 0.962(0.129) & 3.954(0.274) & 0.973(0.091) & 3.562(0.591) & 0.954(0.010) & 2.524(1.274) \\[2pt]
    p = 15 &   &   & \\
QPI &  0.962(0.129) & 3.954(0.274) & 0.962(0.125) & 3.748(0.316) & 0.919(0.209) & 3.493(0.166)\\
PPI &  0.974(0.095) & 4.352(0.370) & 0.960(0.122) & 3.773(0.434) & 0.915(0.213) & 3.498(0.299) \\
PI-KL & 0.948(0.141) & 4.047(0.260) & 0.908(0.173) & 3.483(0.291) & 0.898(0.172) & 2.734(0.787) \\
PI-WA &  0.956(0.133) & 3.884(0.302) & 0.959(0.116) & 3.567(0.422) & 0.998(0.021) & 3.486(1.258)\\
    p = 20 &   &   & \\
QPI & 0.917(0.229) & 3.570(0.338) &   0.948(0.158) & 3.728(0.226) & 0.955(0.147) & 3.683(0.300)  \\
PPI & 0.920(0.214) & 3.615(0.407) &  0.947(0.149) & 3.746(0.340) & 0.951(0.148) & 3.681(0.403) \\
PI-KL & 0.928(0.199) & 3.636(0.213)  & 0.953(0.137) & 3.825(0.308) & 0.890(0.179) & 3.181(0.572) \\
PI-WA & 0.929(0.200) & 3.624(0.149) & 0.956(0.130) & 3.849(0.302) & 0.999(0.009) & 3.604(0.976) \\
   p = 25 &   &   & \\
QPI & 0.896(0.241) & 3.312(0.137) & 0.954(0.147) & 4.069(0.104) & 0.949(0.162) & 3.893(0.164) \\
PPI & 0.897(0.234) & 3.344(0.263) & 0.952(0.139) & 4.086(0.306) & 0.946(0.163) & 3.907(0.311) \\
PI-KL & 0.888(0.219) & 3.481(0.313)  & 0.943(0.146) & 3.844(0.241) & 0.885(0.185) & 3.123(0.586) \\
PI-WA & 0.922(0.212) & 3.654(0.175) & 0.964(0.126) & 3.890(0.338) & 0.961(0.122) & 3.874(0.329) \\
    \bottomrule
    \end{tabular}
\label{Table:fixpvaryingnM3}
\end{table}

\FloatBarrier

\section{\textsc{Proofs}}\label{Appendix:B}

\begin{proof}[\textsc{\textbf{Proof of Proposition 3.1}}]

By \cref{Lemma:variantofnoiseout}, if we take $S(X) = X$ and take $\mathcal{S} = [-M_2, M_2]^d$, we have $Y \stackrel{a.s.}{=} \Xi(X,Z)$; $f$ is measurable. Thus, $\Xi$ could be a satisfied minimizer required by \cref{Eq:empriskDNNTrue}. For this moment, $Z$ is $\text{Uniform}[0,1]$. By \cref{Theorem:Lusin}, there exists a closed set $D \subseteq A$ with $\lambda(A \backslash D)<\epsilon$ such that $\Xi|_D$ is continuous for each $\epsilon>0$. Based on \cref{Theorem:Tietze}, we can further take a continuous functions $\widetilde{\Xi}: A \to \mathbb{R}$, s.t.,
$$
\sup \{|\Xi(x,z)|: (x,z) \in D\}=\sup \{|\widetilde{\Xi}(x,z)|: (x,z) \in A\},
$$
which implies that $|\widetilde{\Xi}(x,z)|$ is also bounded by $M_1$ and $\widetilde{\Xi}(x,z) = \Xi(x,z)$ for all points $(x,z)$ in $D$. Moreover, it is easy to see that:

$$
\lambda(\{(x,z) : \Xi(x,z) \neq \widetilde{\Xi}(x,z)\}) \leq \lambda(A \backslash D)<\epsilon,
$$
which indicates that $\widetilde{\Xi}(x,z)$ would only be different with $\Xi(x,z)$ up to a negligible set. Then, we show that $Z$ could be other distributions without impairing the continuous property of $\widetilde{\Xi}$. 

Take $Z^{\prime}: = (Z^{\prime}_1,\ldots,Z^{\prime}_p) \sim N(0,I_p)$ as an example, let $\Gamma(Z^{\prime}) := \Phi^{-1}(Z^{\prime}_1)$; it is well-known that $\Phi^{-1}(Z^{\prime}_1) \sim \text{Uniform}[0,1]$ by PIT; $\Phi^{-1}(\cdot)$ is the inverse of the CDF of a standard normal random variable; $Z^{\prime}_1$ is the first coordinate of $Z^{\prime}$. Then $\Gamma$ is also continuous. Subsequently, we can consider 
$$
\widetilde{\Xi}(X,Z^{\prime}) = \widetilde{\Xi}(X,\Gamma(Z^{\prime})), 
$$
then $\widetilde{\Xi}(\cdot,\cdot): [-M_2, M_2]^d\times \mathbb{R}^p \to [-M_1,M_1] $ is also continuous. Moreover, $\widetilde{\Xi}(x,z^{\prime}) \neq \Xi(x,z)$ only up to a negligible set. The proof of $Z^{\prime}$ being $\text{Uniform}[0,1]^p$ is similarly. 
\end{proof}

\begin{proof}[\textsc{\textbf{Proof of Theorem 3.1}}]
    The proof is based on the method behind Theorem 1 of \cite{farrell2021deep} and the approximation results of \cite{shen2021deep}. Throughout this proof, $C$ is a constant whose value may change but it does not depend on $n$. Let $A: = [-M_2,M_2]^d \times [0,1]^p$, we first define an optimal function $H_{\text{DNN}}$ which is a DNN in $\mathcal{F}_{\text{DNN}}$:
    \begin{equation*}
       H_{\text{DNN}} := \arg\min_{ \substack{ H_\theta\in\mathcal{F}_{\text{DNN}} \\ \| H_\theta \|_{\infty}\leq M_1}} \| H_\theta - H_0 \|_{L^\infty(A)}.
    \end{equation*}
We can think $H_{\text{DNN}}$ is the oracle optimal DNN estimator of $H_0$ in  $\mathcal{F}_{\text{DNN}}$. Then, we let $\epsilon_n := \| H_{\text{DNN}}- H_0 \|_{L^\infty(A)}$; $\epsilon_n\to 0$ as $n\to\infty$ revealed by \cref{Theorem:approxerrorofc0}. Moreover, under B1, with the choice of $N_1$ and $N_2$ stated in theorem, we can find: 
\begin{equation}\label{Eq:errorbounC0dnn}
\begin{split}
  \| H_{\text{DNN}}- H_0 \|_{L^\infty(A)}  &\leq 19 \sqrt{d} \omega_f^E\left(2 R N_1^{-2 / d} N_2^{-2 / d}\right) \\
  &=  C\cdot (\frac{n^{\frac{d+p}{2(\tau+d+p)}}}{\log n})^{-2/(d+p)}\cdot (\log n)^{-2/(d+p)}\\
  & = C\cdot n^{-\frac{1}{\tau+d+p}};
\end{split}
\end{equation}
here we treat $N_1$ and $N_2$ as $\frac{n^{\frac{d+p}{2(\tau+d+p)}}}{\log n}$ and $\log(n)$, respectively, which does not influence the order of the error bound. 

To give an error bound of $ \left\| \widehat{H}-H_0\right\|_{L(X,Z)}^2$, we first show the loss function in our risk and empirical risk satisfy two conditions which are crucial to do the decomposition of mean square error. Denote the loss function by $\ell(f,Y,X,Z) : = (Y - f(X,Z))^2$. we have:
\begin{equation}\label{Eq:twolossconditions}
    \begin{split}
        |\ell(f, y,x,z)-\ell(g, y,x,z)| & \leq C_{\ell}|f(x,z)-g(x,z)|; \\
        c_1 \mathbb{E}\left[\left(f-H_0\right)^2\right]  \leq \mathbb{E}[\ell(f, Y,X,Z)] & -\mathbb{E}\left[\ell\left(H_0, Y,X,Z\right)\right] \leq c_2 \mathbb{E}\left[\left(f-H_0\right)^2\right];
    \end{split}
\end{equation}
$f$ and $g$ are any two functions map $[-M_2,M_2]^d \times [0,1]^p \to [-M_1,M_1]$; $H_0$ is our desired transformation function; $x, y, z$ are realizations of $X, Y, Z$; $c_1, c_2, C_{\ell}$ are constantly bounded away from 0. For the first row of \cref{Eq:twolossconditions}, we have:
\begin{equation*}
    \begin{split}
         |\ell(f, y,x,z)-\ell(g, y,x,z)| &=  |(y - f(x,z))^2 - (y - g(x,z))^2 | \\
         & = |y^2 + f(x,z)^2 - 2y\cdot f(x,z) - y^2 - g(x,z)^2 + 2y\cdot g(x,z)| \\
         & = |( f(x,z) - g(x,z) )(f(x,z) + g(x,z) - 2y  )|\\
         &\leq C_{\ell}|f(x,z)-g(x,z)|,~C_{\ell} = 4M_1.
    \end{split}
\end{equation*}
For the second row of \cref{Eq:twolossconditions}, we have:
\begin{equation*}
    \begin{split}
\mathbb{E}[\ell(f,Y,X,Z)] -\mathbb{E}\left[\ell\left(H_0, Y,X,Z \right)\right] &= \mathbb{E}[\ell(f,Y,X,Z)] -\mathbb{E}\left[\ell\left(H_0, Y,X,Z \right)\right] \\
& =  \mathbb{E}\big[ f(X,Z)^2 - 2Y\cdot f(X,Z) -  H_0(X,Z)^2 + 2Y\cdot H_0(X,Z)    \big] \\
& =  \mathbb{E}\big[ f(X,Z)^2 - 2H_0(X,Z)\cdot f(X,Z) -  H_0(X,Z)^2 + 2H_0(X,Z)^2    \big] \\
& = \mathbb{E}\big[\left(f(X,Z)-H_0(X,Z)\right)^2\big].\\
    \end{split}
\end{equation*}
Then, we can do the error bound decomposition:
\begin{equation} \label{Eq:errordecom}
    \begin{split}
        & \left\|\widehat{H} - H_0\right\|_{L^2(X,Z)}^2 \\
& \leq \mathbb{E}[\ell(\widehat{H},Y,X,Z)]-\mathbb{E}\left[\ell\left(H_0, Y,X,Z\right)\right]\\
&\leq \mathbb{E}[\ell(\widehat{H},Y,X,Z)]-\mathbb{E}\left[\ell\left(H_0, Y,X,Z\right)\right] - \mathbb{E}_n[\ell(\widehat{H},Y,X,Z)] + \mathbb{E}_n\left[\ell\left(H_{\text{DNN}}, Y,X,Z\right)\right] \\
& = (\mathbb{E} - \mathbb{E}_n) [ \ell(\widehat{H},Y,X,Z) - \ell\left(H_0, Y,X,Z\right)  ] + \mathbb{E}_n [ \ell(H_{\text{DNN}},Y,X,Z) - \ell\left(H_0, Y,X,Z\right)  ];
    \end{split}
\end{equation}
$\mathbb{E}_n$ denotes the sample average, e.g., $\mathbb{E}_n[\ell(\widehat{H},Y,X,Z)] = \frac{1}{n}\sum_{i=1}^n\left( Y_i - \widehat{H}(X_i,Z_i)  \right)^2$
the first inequality is due to the second row of \cref{Eq:twolossconditions}; the second inequality is due to $\widehat{H}$ being the minimizer of empirical risk. Usually,  people call the first term of the r.h.s. of \cref{Eq:errordecom} stochastic error and techniques of the empirical process can bound it; people call the second term approximation error and it can be bounded by Bernstein’s inequality in this case. 

Take a similar procedure to the proof in Theorem 1 of \cite{farrell2021deep}, we can conclude that with probability at least $1-\exp(-\gamma)$, 
\begin{equation}\label{Eq:sqrtrooterrorbound}
    \left\|\widehat{H} - H_0\right\|_{L^2(X,Z)} \leq C\left(\sqrt{\frac{W^2 L^2 \log \left(W^2 L\right)}{n} \log n}+\sqrt{\frac{\log \log n+\gamma}{n}}+\epsilon_n\right) .
\end{equation}
Thus, as long as $\gamma = o(n)$, the r.h.s. of \cref{Eq:sqrtrooterrorbound} converges to 0 as $n\to\infty$ with a probability tends to 1. Furthermore, if we have an additional restriction on the modulus of continuity of $\Xi$, i.e., assuming B1, we can develop a high-probability nonasymptotic error bound. Here, we consider two situations. First, if $(d+p)\left\lfloor N_1^{1 / (d+p)}\right\rfloor$ is larger than $N_1+1$, we have that with probability at least $1-\exp(-\gamma)$:
\begin{equation*}
\begin{split}
    \left\|\widehat{H} - H_0\right\|_{L^2(X,Z)} &\leq C\Bigg(\sqrt{\frac{3^{2(d+p+3)}(d+p)^2n^{\frac{1}{(\tau+d+p)}}(12\log n + 14 + 2(d+p))^2 \log \left(W^2 L\right)   }{n(\log n )^{(2/d+p)}} \log n}   \\
    & +\sqrt{\frac{\log \log n+\gamma}{n}} + C\cdot n^{-\frac{1}{\tau+d+p}} \Bigg)\\
    & = C\Bigg( n^{-\frac{\tau - 1 + d + p}{2(\tau+d+p)}}\mathcal{L}_1(n) + \sqrt{\frac{\log \log n+\gamma}{n}} + n^{-\frac{1}{\tau+d+p}} \Bigg). \\
%    & = C \Bigg( o(n^{-\frac{1}{\tau+d+p}}) +   \sqrt{\frac{\log \log n+\gamma}{n}} + n^{-\frac{1}{\tau+d+p}}    \Bigg)
\end{split}
\end{equation*}
The first inequality is due to \cref{Eq:sqrtrooterrorbound} and \cref{Eq:errorbounC0dnn}; $\mathcal{L}_1(n)$ is a slowly varying function involving a constant and $\log(n)$ terms in the second equation; Then, we can conclude that with probability at least $1-\exp(-\gamma)$: 
\begin{equation*}
\begin{split}
    \left\|\widehat{H} - H_0\right\|_{L^2(X,Z)}^2 &\leq C\Bigg(  n^{-\frac{\tau - 1 + d+p}{\tau+d+p}}\mathcal{L}^2_1(n) + \frac{\log \log n+\gamma}{n} +  n^{-\frac{2}{\tau+d+p}}   \\
    & + 2n^{-\frac{\tau - 1+ d+p}{2(\tau+d+p)}}\mathcal{L}_1(n) \sqrt{\frac{\log \log n+\gamma}{n}} + 2 n^{-\frac{\tau - 1 + d+p}{2(\tau+d+p)}}\mathcal{L}_1(n)  n^{-\frac{1}{\tau+d+p}} + 2\sqrt{\frac{\log \log n+\gamma}{n}} n^{-\frac{1}{\tau+d+p}}   \Bigg) \\
    & \leq C\Bigg(  n^{-\frac{\tau - 1 + d+p}{\tau+d+p}}\mathcal{L}^2_1(n) + \frac{\log \log n+\gamma}{n} +  n^{-\frac{2}{\tau+d+p}}   \\
    & + 2n^{-\frac{\tau - 1+ d+p}{2(\tau+d+p)}}\mathcal{L}_1(n) \frac{\sqrt{\log \log n}+\sqrt{\gamma}}{\sqrt{n}} + 2 n^{-\frac{\tau - 1 + d+p}{2(\tau+d+p)}}\mathcal{L}_1(n)  n^{-\frac{1}{\tau+d+p}} + 2\frac{\sqrt{\log \log n}+\sqrt{\gamma}}{\sqrt{n}} n^{-\frac{1}{\tau+d+p}}   \Bigg).
\end{split}
\end{equation*}
The first inequality comes from extending the square terms; the second inequality relies on the fact $\sqrt{a +b} \leq \sqrt{a} + \sqrt{b}$. We take $\gamma = n^{\frac{d+p-1}{\tau+d+p}}$, then we will have below results with probability at least $1-\exp(-n^{\frac{d+p-1}{\tau+d+p}})$:
\begin{equation}
   \left\|\widehat{H} - H_0\right\|_{L^2(X,Z)}^2 \leq 
  C\cdot n^{-\frac{2}{\tau+d+p}} +  o(n^{-\frac{2}{\tau+d+p}});~ \text{for}~ d + p \geq 2; \tau>1; (d+p)\left\lfloor N_1^{1 / (d+p)}\right\rfloor \geq N_1+1;
\end{equation}

Second, if $N_1+1$ is larger than $(d+p)\left\lfloor N_1^{1 / (d+p)}\right\rfloor$, we can do the similar analysis as above. It is not hard to find that with probability at least $1-\exp(-\gamma)$:
\begin{equation*}
\begin{split}
    \left\|\widehat{H} - H_0\right\|_{L^2(X,Z)} &\leq C\Bigg( n^{-\frac{\tau}{2(\tau+d+p)}}\mathcal{L}_2(n) +   \sqrt{\frac{\log \log n+\gamma}{n}} + n^{-\frac{1}{\tau+d+p}} \Bigg);
\end{split}
\end{equation*}
where $\mathcal{L}_2(n)$ is a slowly varying function involving a constant and
log(n) terms. Then, we can take $\gamma = n^{\frac{d+p}{\tau+d+p}}$ and find with probability at least $1-\exp(-n^{\frac{d+p}{\tau+d+p}})$:
\begin{equation}
   \left\|\widehat{H} - H_0\right\|_{L^2(X,Z)}^2 \leq C\cdot n^{-\frac{2}{\tau+d+p}} + o(n^{-\frac{2}{\tau+d+p}});~\text{for}~d+p\geq 2; \tau>2;(d+p)\left\lfloor N_1^{1 / (d+p)}\right\rfloor < N_1+1;
\end{equation}

\end{proof}

\begin{proof}[\textsc{\textbf{Proof of Theorem 3.2}}]
For convenience, we present the convergence result in \cref{Theorem:conditionalDFcon} here:
\begin{equation*}
\sup_{y} \Big|\widehat{F}_{\widehat{H}(x_f,Z)}(y) -   F_{Y|x_f}(y)\Big| \xrightarrow{p} 0,~\text{as}~n\to \infty, S\to \infty;
\end{equation*}
By the triangle inequality, we have:
\begin{equation}\label{Eq:T31proofdecom}
    \begin{split}
&\sup_{y} \Big|\widehat{F}_{\widehat{H}(x_f,Z)}(y) -   F_{Y|x_f}(y)\Big| \\
& = \sup_{y} \Big|\widehat{F}_{\widehat{H}(x_f,Z)}(y) - \widehat{F}_{H_0(x_f,Z)}(y) + \widehat{F}_{H_0(x_f,Z)}(y) - F_{H_0(x_f,Z)}(y) + F_{H_0(x_f,Z)}(y)  -   F_{Y|x_f}(y)\Big| \\
& \leq \sup_{y} \Big| \widehat{F}_{\widehat{H}(x_f,Z)}(y) - \widehat{F}_{H_0(x_f,Z)}(y)  \Big|  + \sup_{y} \Big| \widehat{F}_{H_0(x_f,Z)}(y) - F_{H_0(x_f,Z)}(y)   \Big|  +  \sup_{y} \Big| F_{H_0(x_f,Z)}(y)  -   F_{Y|x_f}(y)  \Big|;
    \end{split}
\end{equation}
$\widehat{F}_{H_0(x_f,Z)}(y): = \text{the empirical distribution of }~\{H_0(x_f,Z_i)\}_{i=1}^S$; $F_{H_0(x_f,Z)}(y): =$ the distribution of $H_0(x_f,Z)$. For the middle term on the r.h.s. of \cref{Eq:T31proofdecom}, it converges to 0 a.s. due to Glivenko–Cantelli theorem. For the last term, it equals 0 due to the definition of $H_0(x_f,Z)$. Thus, we focus on the first term on the r.h.s. of \cref{Eq:T31proofdecom}.

First, we define
\begin{equation*}
    \text{if}~\Delta_i(y) = \begin{cases}
    1, & H_0(x_f,Z_i)\leq y\\
    0, & H_0(x_f,Z_i)> y
    \end{cases};
\end{equation*}
then, we can get:
\begin{equation}\label{Eq55}
    \widehat{F}_{\widehat{H}(x_f,Z)}(y) = \frac{1}{S}\sum_{i=1}^{S}\Delta_i(y+ H_0(x_f,Z_i) - \widehat{H}(x_f,Z_i)); \widehat{F}_{H_0(x_f,Z)}(y) = \frac{1}{S}\sum_{i=1}^{S}\Delta_{i}(y).
\end{equation}
For handling the randomness of $H_0(x_f,Z_i) - \widehat{H}(x_f,Z_i)$ inside $\Delta_{i}(\cdot)$ of \cref{Eq55}, we use nonrandom $\eta_{S},S=1,2,3,\ldots,$ to replace $H_0(x_f,Z_i) - \widehat{H}(x_f,Z_i)$. Then, we can consider the process:
\begin{equation*}
    r_{S}(y,\eta_{S}) := \widehat{F}_{\widehat{H}(x_f,Z)}(y) - \widehat{F}_{H_0(x_f,Z)}(y) = \frac{1}{S}\sum_{i=1}^{S}\left( \Delta_{i}(y+\eta_{S}) - \Delta_{i}(y) \right).
\end{equation*}
Indeed, we have:
\begin{equation*}
    \mathbb{P}(\sup_{y}|\widehat{F}_{\widehat{H}(x_f,Z)}(y) - \widehat{F}_{H_0(x_f,Z)}(y)|>\epsilon)\leq \mathbb{P}(\sup_{y}\sup_{|\eta_S|\leq S^{-\lambda_1}}|r_{S}(y,\eta_{S})|>\epsilon) + \mathbb{P}(|H_0(x_f,Z) - \widehat{H}(x_f,Z)|>S^{-\lambda_1}).
\end{equation*}
Without loss of generality, we select an appropriate $\lambda_1$ to make sure the second term on the right-hand side of the above inequality converges to 0 with a high probability. This is reasonable since we can show $|H_0(x_f,Z) - \widehat{H}(x_f,Z)|\xrightarrow{p} 0$ with probability tending to 1 as $n\to\infty$ by \cref{Theorem:errorboundofC0f}; see more details from the proof of Theorem 4.1.

Then, we shall show $\mathbb{P}(\sup_{y}\sup_{|\eta_S|\leq S^{-\lambda_1}}|r_{S}(y,\eta_{S})|>\epsilon)$ also converges to 0. Since this term depends on the continuum of values of $y$, we can partition the real axis into $N_{S} \sim S^{\lambda_2}$ parts by points:
\begin{equation*}
    -\infty = y_{0}<y_{1}<\cdots y_{k}< \cdots<y_{N_{S}-1}<y_{N_S} = \infty,~\text{where}~F_{Y|x_f}(y_k) = kN_{S}^{-1}; \lambda_2 > 0.
\end{equation*}
Hence, for $y_{r}$ and $y_{r+1}$ such that $y_{r}\leq y \leq y_{r+1}$, we have:
\begin{equation*}
    y_{r} + \eta_{S} \leq y + \eta_{S} \leq y_{r+1} + \eta_{S}.
\end{equation*}
In addition, since $\Delta_i(y)$ is monotonic, we obtain:
\begin{equation*}
\begin{split}
     r_{S}(y,\eta_{S}) &\geq r_{S}(y_r,\eta_{S}) + \frac{1}{S}\sum_{i=1}^{S}\Delta_i(y_r) - \frac{1}{S}\sum_{i=1}^{S}\Delta_i(y_{r+1}); \\
     r_{S}(y,\eta_{S}) &\leq r_{S}(y_{r+1},\eta_{S}) + \frac{1}{S}\sum_{i=1}^{S}\Delta_i(y_{r+1}) - \frac{1}{S}\sum_{i=1}^{S}\Delta_i(y_{r}).
\end{split}
\end{equation*}
Therefore, we have:
\begin{equation}\label{Eq61}
\begin{split}
    \sup_{y}\sup_{|\eta_S|\leq S^{-\lambda_1}}&|r_{S}(y,\eta_{S})|\\ 
    &\leq \sup_{k \leq N_{S}-1}\sup_{|\eta_S|\leq S^{-\lambda_1}}|r_{S}(y_{k+1},\eta_{S})| \\
    & + \sup_{k \leq N_{S}}\sup_{|\eta_S|\leq S^{-\lambda_1}}|r_{S}(y_{k},\eta_{S})|\\
    & + \sup_{|t_1 - t_2|\leq N_{S}^{-1}}\frac{1}{S}\left |\sum_{i=1}^{S}\left(\Delta_{i}(F_{Y|x_f}^{-1}(t_1)) -  \Delta_{i}(F_{Y|x_f}^{-1}(t_2)) \right)\right |.
\end{split}
\end{equation}
For the last term on the r.h.s. of \cref{Eq61}:
\begin{equation*}
\begin{split}
    &\sup_{|t_1 - t_2|\leq N_{S}^{-1}}\frac{1}{S}\left |\sum_{i=1}^{S}\left(\Delta_{i}(F_{Y|x_f}^{-1}(t_1)) -  \Delta_{i}(F_{Y|x_f}^{-1}(t_2)) \right)\right | \\
    &= \sup_{|t_1 - t_2|\leq N_{S}^{-1}}\frac{1}{S}\left |\sum_{i=1}^{S}\left(\Delta_{i}(F_{Y|x_f}^{-1}(t_1)) - t_1 +t_1 -t_2 + t_2 -  \Delta_{i}(F_{Y|x_f}^{-1}(t_2)) \right)\right |\\
    & \leq \sup_{t_1, s.t. |t_1 - t_2|\leq N_{S}^{-1}}\left|\frac{1}{S}\sum_{i=1}^{S}\Delta_{i}(F_{Y|x_f}^{-1}(t_1)) - t_1 \right| + \sup_{|t_1 - t_2|\leq N_{S}^{-1}}\left| t_1 -t_2  \right| +  \sup_{t_2, s.t.|t_1 - t_2|\leq N_{S}^{-1}}\left| t_2  - \frac{1}{S}\sum_{i=1}^{S} \Delta_{i}(F_{Y|x_f}^{-1}(t_2))    \right|. 
\end{split}
\end{equation*}
By the Glivenko–Cantelli theorem, it is obvious that $\sup_{|t_1 - t_2|\leq N_{S}^{-1}}\frac{1}{S}\left |\sum_{i=1}^{S}\left(\Delta_{i}(F_{Y|x_f}^{-1}(t_1)) -  \Delta_{i}(F_{Y|x_f}^{-1}(t_2)) \right)\right |$ is $o_p(1)$. Next, we consider the second term of the r.h.s of \cref{Eq61}:
\begin{equation}\label{Eq63}
\begin{split}
    &\sup_{k \leq N_{S}}\sup_{|\eta_S|\leq S^{-\lambda_1}}|r_{S}(y_{k},\eta_{S})|\\
    &=  \sup_{k \leq N_{S}}\sup_{|\eta_S|\leq S^{-\lambda_1}}\left|  \frac{1}{S}\sum_{i=1}^{S}\left( \Delta_{i}(y_{k}+\eta_{S}) - \Delta_{i}(y_k) \right) \right| \\
    &=  \sup_{k \leq N_{S}}\sup_{|\eta_S|\leq S^{-\lambda_1}}\left|  \frac{1}{S}\sum_{i=1}^{S} \Delta_{i}(y_{k}+\eta_{S}) - F_{Y|x_f}(y_{k}+\eta_{S}) + F_{Y|x_f}(y_{k}+\eta_{S}) - F_{Y|x_f}(y_{k}) +  F_{Y|x_f}(y_{k}) -  \frac{1}{S}\sum_{i=1}^{S} \Delta_{i}(y_{k})    \right|\\
    &\leq \sup_{k \leq N_{S}}\sup_{|\eta_S|\leq S^{-\lambda_1}}\Bigg\{  \left|  \frac{1}{S}\sum_{i=1}^{S} \Delta_{i}(y_{k}+\eta_{S}) - F_{Y|x_f}(y_{k}+\eta_{S}) \right| +\left| F_{Y|x_f}(y_{k}+\eta_{S}) - F_{Y|x_f}(y_{k}) \right| \\
    &+ \left| F_{Y|x_f}(y_{k}) -  \frac{1}{S}\sum_{i=1}^{S} \Delta_{i}(y_{k})  \right| \Bigg\}.
\end{split}
\end{equation}
Applying the Glivenko–Cantelli theorem again, we can find the first and third term in the r.h.s. of \cref{Eq63} converges to 0 a.s.. For the middle term:
\begin{equation*}
\begin{split}
    &\sup_{k \leq N_{S}}\sup_{|\eta_S|\leq S^{-\lambda_1}}\left| F_{Y|x_f}(y_{k}+\eta_{S}) - F_{Y|x_f}(y_{k}) \right|\\
    &=  \sup_{k \leq N_{S}}\sup_{|\eta_S|\leq S^{-\lambda_1}}\left| F_{Y|x_f}(y_k) + F_{Y|x_f}^{\prime}(\Tilde{y})\eta_S - F_{Y|x_f}(y_k)  \right| \\
    & \leq \sup_{y}\sup_{|\eta_S|\leq S^{-\lambda_1}}\left|F_{Y|x_f}^{\prime}(\Tilde{y})\eta_S  \right| \overset{}{\to} 0,~\text{Under A1.}
\end{split}
\end{equation*}
We can do a similar analysis for the first term of the r.h.s of \cref{Eq61}. Combining all the parts together, we finish the proof.

\end{proof}

\begin{proof}[\textsc{\textbf{Proof of Theorem 4.1}}]
For condition (1), we can write:
\begin{equation*}
    \begin{split}
        R_f & = Y_f - \Psi(F_{Y|x_f}) + \Psi(F_{Y|x_f}) -  \widehat{Y}_{f,L_2}, \\
        R_f^* &= Y^*_f - \Psi(\widetilde{F}_{Y|x_f}) + \Psi(\widetilde{F}_{Y|x_f}) -  \widehat{Y}^*_{f,L_2};
    \end{split}
\end{equation*}
$\Psi(F_{Y|x_f})$ and $\Psi(\widetilde{F}_{Y|x_f})$ are some functional of $F_{Y|x_f}$ and $\widetilde{F}_{Y|x_f}$, respectively; $\widetilde{F}_{Y|x_f}$ represents the CDF of $\widehat{H}(X_f,Z)$ and we approximate it by the empirical distribution of $\{\widehat{H}(X_f, Z_j)\}_{j=1}^S$. By the Glivenko–Cantelli theorem, the empirical distribution of $\{\widehat{H}(X_f, Z_j)\}_{j=1}^S$ converges to $\widetilde{F}_{Y|x_f}$ uniformly almost sure with $S$ converges to infinity in an appropriate rate; see more details about the rate later. Thus, we take $\widetilde{F}_{Y|x_f}$ to do theoretical analysis. We have:
\begin{equation*}
    \begin{split}
        \epsilon_f = Y_f - \Psi(F_{Y|x_f})~&;~\epsilon_f^* = Y^*_f - \Psi(\widetilde{F}_{Y|x_f});\\
         e_f = \Psi(F_{Y|x_f}) -  \widehat{Y}_{f,L_2}~&;~e_f^* = \Psi(\widetilde{F}_{Y|x_f}) -  \widehat{Y}^*_{f,L_2}.
    \end{split}
\end{equation*}
Since we proposed the predictive interval which is centered around the optimal $L_2$ point prediction, we can take $\Psi(F_{Y|x_f}) = \mathbb{E}(Y|x_f)$ and  $\Psi(\widetilde{F}_{Y|x_f}) = \mathbb{E}(\widehat{H}(x_f,Z))$ conditional on observations $\{X_i,Y_i,Z_i\}_{i=1}^n$ and $x_f$; we use $\widehat{Y}_{f,L_2} = \frac{1}{S}\sum_{j=1}^S \widehat{H}(X_f, Z_j)$ to approximate $\Psi(\widetilde{F}_{Y|x_f})$.

For condition (2), we have
\begin{equation}\label{Eq:condi2}
    \begin{split}
        &\sup _x\left|\mathbb{P}^*\left(\epsilon_f^* \leq x\right)-\mathbb{P}\left(\epsilon_f \leq x\right)\right| \\
         = & \sup _x\left|\mathbb{P}^*\left(Y^*_f - \Psi(\widetilde{F}_{Y|x_f}) \leq x\right)-\mathbb{P}\left( Y_f - \Psi(F_{Y|x_f}) \leq x\right)\right| \\
         = & \sup _x\left|\widetilde{F}_{Y|x_f}\left(\Psi(\widetilde{F}_{Y|x_f}) + x\right)- F_{Y|x_f}\left(\Psi(F_{Y|x_f}) + x\right)\right|\\
         \leq & \sup _x\left|\widetilde{F}_{Y|x_f}\left(\Psi(\widetilde{F}_{Y|x_f}) + x\right)- F_{Y|x_f}\left(\Psi(\widetilde{F}_{Y|x_f}) + x\right)\right| \\
         + & \sup _x\left|F_{Y|x_f}\left(\Psi(\widetilde{F}_{Y|x_f}) + x\right)- F_{Y|x_f}\left(\Psi(F_{Y|x_f}) + x\right)\right|.
    \end{split}
\end{equation}
Thus, the first term on the r.h.s. of \cref{Eq:condi2} converges to 0 in probability due to \cref{Theorem:conditionalDFcon} and Glivenko–Cantelli theorem. For the second term on the r.h.s. of \cref{Eq:condi2}, since $F_{Y|x_f}$ is $C^1$ due to A1, we have:
\begin{equation}\label{Eq:2ndtermcondi2}
    \begin{split}
        & \sup _x\left|F_{Y|x_f}\left(\Psi(\widetilde{F}_{Y|x_f}) + x\right)- F_{Y|x_f}\left(\Psi(F_{Y|x_f}) + x\right)\right| \\
        = & \sup _{x,\Tilde{x}}\left|F_{Y|x_f}\left(\Psi(F_{Y|x_f}) + x\right) + f_{Y|x_f}\left(\Tilde{x}\right)(\Psi(\widetilde{F}_{Y|x_f}) - \Psi(F_{Y|x_f}))  - F_{Y|x_f}\left(\Psi(F_{Y|x_f}) + x\right)\right|.
    \end{split}
\end{equation}
By the fact that the domain $\mathcal{Y}$ is compact and $\Psi(\widetilde{F}_{Y|x_f}) \xrightarrow{p} \Psi(F_{Y|x_f})$ as $n\to\infty$ and $S\to\infty$, we have \cref{Eq:2ndtermcondi2} converges to 0 in probability. 

For condition (3), we first analyze $e_f$ term. We take $\widehat{Y}_{f,L_2} = \frac{1}{S}\sum_{j=1}^S \widehat{H}(x_f, Z_j)$. Since $\{Z_j\}_{j=1}^S$ are independent, we have $\mathbb{E}_{Z}\left( \frac{1}{S}\sum_{j=1}^S \widehat{H}(x_f, Z_j) \right) = \mathbb{E}_{Z}\left( \widehat{H}(x_f, Z)  \right)$ conditional on $\{X_i,Y_i,Z_i\}_{i=1}^n$ and $x_f$.
% Since $L_2$ convergence implies $L_1$ convergence, we have:
% \begin{equation}
%     \mathbb{E}_{X,Z}\left[ \Big| \widehat{H}_\theta(X,Z)- H_0(X,Z)\Big| \right] \to 0
% \end{equation}
%with a rate $O(n^{\frac{-\xi}{\xi + d}})$ and \textit{high probability}. 
Later, we try to analyze the weak convergence of $\widehat{H}(X_f,Z)$ conditional on $x_f$. Since $X$ and $Z$ are independent, we observe that
\begin{equation}\label{Eq:condiH}
  \mathbb{E}_{X,Z}\left[ \left( \widehat{H}(X,Z)- H_0(X,Z)\right)^2 \right] = \mathbb{E}_{X}\left[ \mathbb{E}_{Z} \left( \left( \widehat{H}(X,Z)- H_0(X,Z)\right)^2 \bigg| X = x_f \right) \right].
\end{equation}
Due to \cref{Theorem:errorboundofC0f}, with probability tending to 1, \cref{Eq:condiH} converges to 0. Since we assume that the density of $X$ is positive for all points in the domain $\mathcal{X}$ in A1, \cref{Eq:condiH} converges to 0 if and only if 
\begin{equation*}
    \mathbb{E}_{Z} \left( \left( \widehat{H}(X,Z)- H_0(X,Z)\right)^2 \bigg| X = x_f \right) \to 0,
\end{equation*}
for any point $X_f = x_f$. By the Markov inequality, we further have:
 \begin{equation*}
     \widehat{H}(x_f,Z)- H_0(x_f,Z) \xrightarrow{p} 0. 
 \end{equation*}
In addition, due to the fact $L_2$ convergence implies $L_1$ convergence, we further have
\begin{equation*}
     \mathbb{E}_{Z} \left( \Big| \widehat{H}(x_f,Z)- H_0(x_f,Z)\Big| \right) \to 0.
\end{equation*}
All in all,
\begin{equation*}
     \mathbb{E}_{Z}\left( \widehat{H}(x_f, Z)  \right) -  \mathbb{E}_{Z}\Big(  H_{0}(X_f, Z)  \Big) \leq \mathbb{E}_Z\left( \Big| \widehat{H}(x_f,Z)- H_0(x_f,Z)\Big| \right) \to 0.
\end{equation*}
The above result is also conditional on $\{X_i,Y_i,Z_i\}_{i=1}^n$. Applying the iterated expectation again, this convergence result still holds for unconditional cases. Note that $\mathbb{E}_{Z}(H_{0}(x_f, Z)) = \Psi(F_{Y|x_f})$. Thus, with probability tending to 1, we can write:
\begin{equation*}
    \mathbb{E}_{Z}\left( \widehat{H}(x_f, Z)  \right) =  \Psi(F_{Y|x_f}) + O(1/\tau_n). 
\end{equation*}
For some appropriate sequence $\tau_n \to \infty$. On the other hand,  by the CLT, we have
\begin{equation*}
    a_S(\widehat{Y}_{f,L_2} - \mathbb{E}_{Z}(\widehat{H}(x_f, Z))) \to N(0,\sigma_\infty);
\end{equation*}
$\sigma_\infty < \infty$ since $\widehat{H}$ is bounded by assumption. If we take an appropriate sequence of $a_S$ such that $a_S\cdot O_p(1/\tau_n) = o_p(1)$, we can finally get:
\begin{equation*}
\begin{split}
    & a_S\left(\widehat{Y}_{f,L_2} - \mathbb{E}_{Z}(\widehat{H}(X_f, Z))\right) \\
    = &  a_S\left(\widehat{Y}_{f,L_2} - \Psi(F_{Y|x_f}) + O_p(1/\tau_n)\right)\\
    = & a_S\left(\widehat{Y}_{f,L_2} - \Psi(F_{Y|x_f})\right) + o_p(1).
\end{split}
\end{equation*}
Since $a_S$ depends on $n$, we change the notation to $a_{n}$. Then, the above equations imply that:
\begin{equation*}
    a_{n}e_f \xrightarrow{d} N(0,\sigma_{\infty}).
\end{equation*}
To analyze $a_{n}e^*_f$ in the bootstrap world, we first notice that $\widehat{H}(x_f,Z)- H_0(x_f,Z) \xrightarrow{p} 0$ with probability tending to 1. In other words, there is an appropriate sequence of sets $\Omega_{n}$, such that $\mathbb{P}\left((\{X_i,Y_i,Z_i\}_{i=1}^n)\notin\Omega_{n}\right) = o(1)$ and $\widehat{H}(x_f,Z)- H_0(x_f,Z) \xrightarrow{p} 0$ for $\{X_i,Y_i,Z_i\}_{i=1}^n\in\Omega_{n}$. Focus on such $\{X_i,Y_i,Z_i\}_{i=1}^n$, we can get $a_{n}e^*_f \xrightarrow{d^*} N(0,\sigma_{\infty})$ in probability since $\widehat{H}(x_f,Z)- H_0(x_f,Z) \xrightarrow{p} 0$ and $\widehat{H}^*(x_f,Z)- \widehat{H}(x_f,Z) \xrightarrow{p^*} 0$. By Polya’s theorem, we show:
\begin{equation*}
    \sup _x\left|\mathbb{P}^*\left(a_{n} e_f^* \leq x\right)-\mathbb{P}\left(a_{n} e_f \leq x\right)\right| \xrightarrow{p} 0 \text{~as~} n \to \infty.
\end{equation*}
%{\color{red} It seems we can not let $S$ to be too large since we need $\tau_S\cdot O_p(1/\tau_n) = o_p(1)$? In practice, a statement should be $\tau_{S,n} \to \infty$ under $\tau_n \to \infty$.}

For condition (4), the randomness of $\epsilon_f$ only depends on $Y_f$; the randomness of $e_f$ depends on $\{X_i,Y_i,Z_i\}_{i=1}^n$ which are independent with $Y_f$; thus $\epsilon_f \indep e_f$. For $\epsilon^*_f$ in the bootstrap world, the randomness of $\epsilon_f$ only depends on $Y^*_f$; the randomness of $e^*_f$ depends $\{Y^*_i,Z^*_i\}_{i=1}^n$ which are independent with $Y^*_f$; thus $\epsilon^*_f \indep e^*_f$. 

Under A5, \cref{Eq:lessdatareason} is straightforward with the definition of a convolution between the empirical distribution and standard normal distribution and the Glivenko–Cantelli theorem. 
\end{proof}

\begin{proof}[\textsc{\textbf{Proof of Lemma 5.1}}]
We give the proof for the DLMF prediction approach. This lemma also holds for deep generator and benchmark prediction approaches. Conditional on $\{(X_i,Y_i,Z_i)\}_{i=1}^n$, we have
\begin{equation}\label{Eq:lemma51proofweakcon}
     \overline{L}_j \xrightarrow{p} \mathbb{E}\left( \left[Y_{j,L_2} - \mathbb{E}\Big(\widehat{H} (X_{j},Z)|X_j,\{(X_i,Y_i,Z_i)\}_{i=1}^n\Big)\right]^2 \bigg| X_j = x_j  \right),~\text{as}~S\to\infty~\text{and}~R\to\infty. 
\end{equation}
By the conditional Jensen’s inequality, we have
\begin{equation}\label{Eq:lemma51proofcje}
\begin{split}
    &\mathbb{E}\left( \left[Y_{j,L_2} - \mathbb{E}(\widehat{H} (X_{j},\eta)|X_j,\{(X_i,Y_i,Z_i)\}_{i=1}^n)\right]^2 \bigg| X_j =x_j  \right) \\
    &> \left( \mathbb{E}\left( Y_{j,L_2} - \mathbb{E}(\widehat{H} (X_{j},\eta)|X_j,\{(X_i,Y_i,Z_i)\}_{i=1}^n) \bigg| X_j = x_j \right)  \right)^2\\
    & = \left(  Y_{j,L_2} - \mathbb{E}(\widehat{H} (X_{j},\eta)|X_j = x_j )   \right)^2.
\end{split}
\end{equation}
The last equality is due to the tower property of conditional expectation. Denote $\mathbb{E}\Big( \Big[Y_{j,L_2} - $ $\mathbb{E}(\widehat{H} (X_{j},\eta)|X_j,\{(X_i,Y_i,Z_i)\}_{i=1}^n)\Big]^2 \bigg| X_j =x_j  \Big)$ by $\widetilde{E}$, \cref{Eq:lemma51proofcje} implies that there is a region such that $\exists \epsilon > 0, \forall T \in [\widetilde{E} - \epsilon, \widetilde{E} + \epsilon]$, $T > \left(  Y_{j,L_2} - \mathbb{E}(\widehat{H} (X_{j},\eta)|X_j = x_j )   \right)^2$. On the other hand, \cref{Eq:lemma51proofweakcon} implies $\forall \varepsilon > 0$, $\mathbb{P}\Big(|\overline{L}_j  - \widetilde{E}| \geq \varepsilon\Big) \to 0$. Then, it is easy to see the statement in \cref{Lemma:upperboundofbarLj}.
\end{proof}

\begin{Theorem}[Lusin's Theorem according to our notations]\label{Theorem:Lusin}
    Given a measurable set $A: = [-M_2, M_2]^d\times \mathcal{U}$ and given measurable $\Xi: A \to \mathbb{R}$. Then, for each $\epsilon>0$, there exists a closed set $D \subseteq A$ with $\lambda(A \backslash D)<\epsilon$ such that $\Xi|_D$ is continuous; $\Xi|_D$ means the restriction of $\Xi$ to $D$. 
\end{Theorem}
See the standard Lusin's Theorem from \cite{wage1975generalization} for reference. We should mention that $\Xi|_D$ is continuous and is not equivalent to $\Xi$ being continuous at each point of $D$. To see the reasoning, we can consider the characteristic function of the rationals which is not continuous at any point, but it will be continuous after removing a measure-0 set.

\begin{Theorem}[Tietze Extension Theorem according to our notations]\label{Theorem:Tietze}
Let $A$ be a normal space, let $D \subseteq A$ be a closed subspace, and let $\Xi: D \to \mathbb{R}$ be a continuous function. There exists a continuous function $\widetilde{\Xi}: A \to \mathbb{R}$ such that $\widetilde{\Xi} = \Xi$ for all points in $D$.
\end{Theorem}

The Tietze Extension Theorem is also known as the Urysohn–Brouwer–Tietze lemma.

\begin{Theorem}[Theorem 4.3 of \cite{shen2021deep}]\label{Theorem:approxerrorofc0}
     Let $f_{*}$ be a uniformly continuous function defined on $E \subseteq[-R, R]^d$. For arbitrary $N_1 \in \mathbb{N}^{+}$and $N_2 \in \mathbb{N}^{+}$, there exists a function $f_{\text{DNN}}$ implemented by a ReLU DNN with width $3^{d+3} \max \left\{d\left\lfloor N_1^{1 / d}\right\rfloor, N_1+1\right\}$ and depth $12 N_2+14+2 d$ such that
$$
\|f_{*}-f_{\text{DNN}}\|_{L^{\infty}(E)} \leq 19 \sqrt{d} \omega_f^E\left(2 R N_1^{-2 / d} N_2^{-2 / d}\right);
$$
$\omega_f^E(r)$ is the so-called modulus of continuity of $f$ on a subset $E$ belongs to the input space $S$:
$$
\omega_f^E(r):=\sup \left\{\left|f\left(x_1\right)-f\left(x_2\right)\right|; d_S\left(x_1, x_2\right) \leq r, x_1, x_2 \in E\right\},~\text{for any }~ r \geq 0.
$$
$d_S(\cdot,\cdot)$ is a metric define in $S$. 

\end{Theorem}

\end{appendices}

\clearpage
\bibliographystyle{apalike}
\bibliography{refs}
\end{document}